\pdfoutput=1

\documentclass[11pt]{article}

\usepackage[dvipsnames]{xcolor, colortbl}
\usepackage{ACL2023}

\usepackage{times}
\usepackage{latexsym}
\usepackage{enumitem}
\usepackage{array,multirow}
\usepackage{float}
\usepackage{comment}

\usepackage[T1]{fontenc}

\usepackage[utf8]{inputenc}

\usepackage{microtype}

\usepackage{graphicx}
\usepackage{amsmath}
\usepackage{mathrsfs}
\usepackage{url}
\usepackage{booktabs,siunitx}

\usepackage{dsfont}
\newcommand{\mc}[3]{\multicolumn{#1}{#2}{#3}}

\usepackage{inconsolata}
\usepackage{amsfonts}

\makeatletter
\renewcommand{\paragraph}{%
  \@startsection{paragraph}{4}%
  {\z@}{1.0ex plus 0.2ex minus .2ex}{-1em}%
  {\normalsize \bf }%
}
\makeatother

\title{Training Language Models to Generate Text with Citations via Fine-grained Rewards}

\author{Chengyu Huang \\
  National University of Singapore \\
  \texttt{e0376956@nus.edu.sg} \\\And
  Zeqiu Wu \\
  University of Washington \\
  \texttt{zeqiuwu1@uw.edu} \\\AND
  Yushi Hu \\
  University of Washington \\
  \texttt{yushihu@uw.edu} \\\And
  Wenya Wang \\
  Nanyang Technological University \\
  \texttt{wangwy@ntu.edu.sg} \\}

\usepackage{hyperref}

\begin{document}
\maketitle
\begin{abstract}
While recent Large Language Models (LLMs) have proven useful in answering user queries, they are prone to hallucination, and their responses often lack credibility due to missing references to reliable sources. An intuitive solution to these issues would be to include in-text citations referring to external documents as evidence. While previous works have directly prompted LLMs to generate in-text citations, their performances are far from satisfactory, especially when it comes to smaller LLMs. In this work, we propose an effective training framework using \emph{fine-grained rewards} to teach LLMs to generate highly supportive and relevant citations, while ensuring the correctness of their responses. We also conduct a systematic analysis of applying these fine-grained rewards to common LLM training strategies, demonstrating its advantage over conventional practices. 
We conduct extensive experiments on Question Answering (QA) datasets taken from the ALCE benchmark and validate the model's generalizability using EXPERTQA. On LLaMA-2-7B, the incorporation of fine-grained rewards achieves the best performance among the baselines, even surpassing that of GPT-3.5-turbo.\footnote{Our code and data is available at \url{https://github.com/HCY123902/atg-w-fg-rw}.}
\end{abstract}

\section{Introduction}

Despite the recent success of Large Language Models (LLMs), their outputs often contain incorrect or unverifiable claims \cite{ji2022survey}. 
This reduces the usefulness of their generated content, and more importantly, hurts the user trust in LLMs.
Previous work \cite{guu2020retrieval, lewis2020rag, izacard2022atlas} attempts to address this issue with retrieval-augmented generation (i.e., RAG), which makes the LLM condition its generation on information retrieved from an external knowledge source for the task input.
However, RAG cannot fully solve the issue because there is no guarantee that the retrieved knowledge is relevant nor that the LM-generated output is always consistent with the knowledge. 

\begin{figure}[t!]
    \includegraphics[width=.48\textwidth]{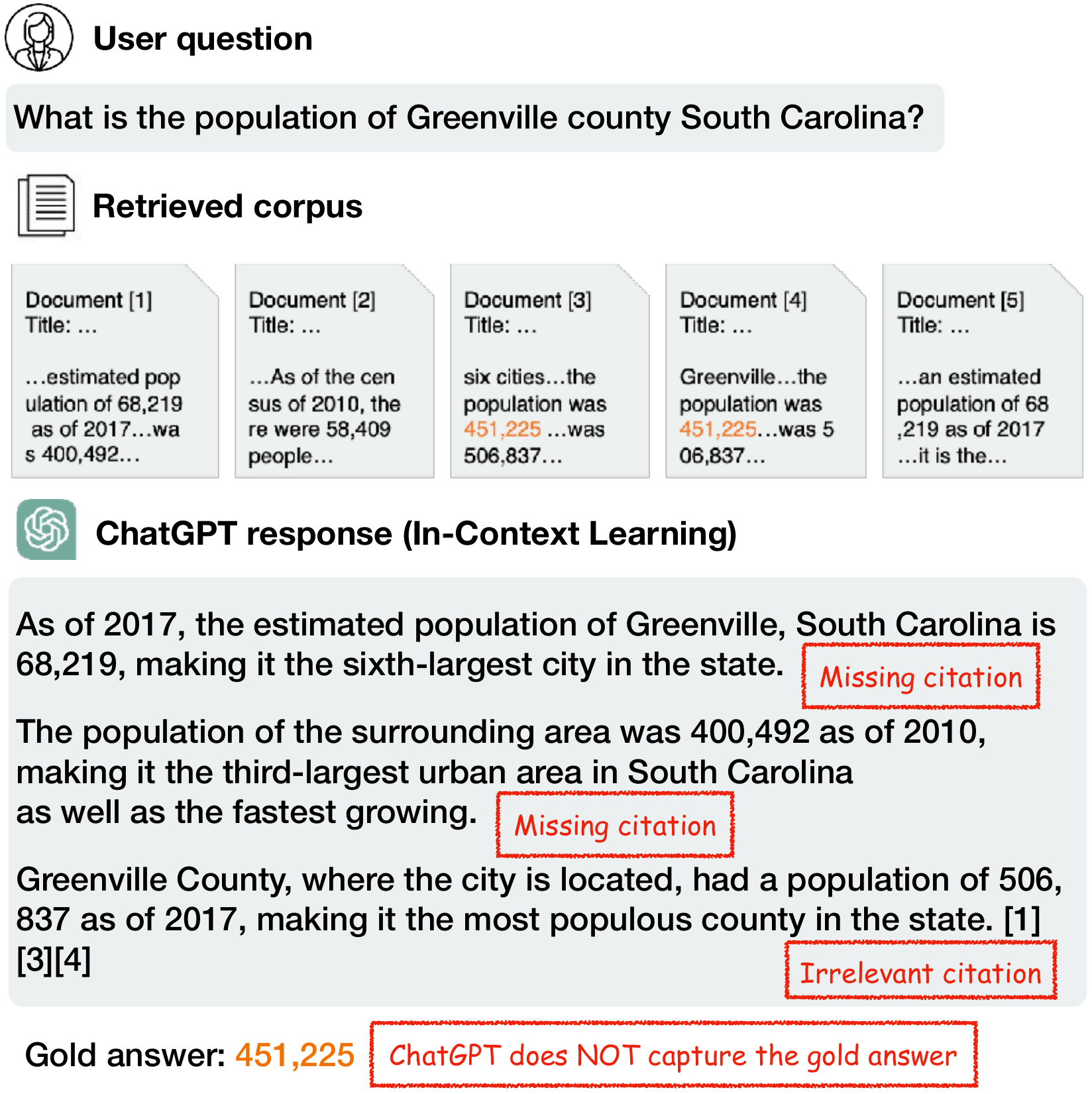}
    \caption{\label{fig:reward}An example of ChatGPT performing the task of attributable generation. The model takes a question, retrieved passages, and the task instruction (omitted due to space limit) as the input, and generates a response with in-text citations. The response has 3 sentences, 2 of which 
    do not have supporting citations. The third one 
    has an irrelevant citation [1]. Moreover, ChatGPT does not capture the correct answer
    (\textcolor{Orange}{451,225}) 
    mentioned in passages [3] and [4].
    }
\end{figure}

Recent studies \cite{nakano2021webgpt, gao2023enabling} propose to have LLMs generate answers with in-text citations that provide evidence for the generated content, often referred to as \textit{attributable} text generation. The cited references provide easy access for end users to perform fact-checking on the model generations. 
Most existing work \cite{malaviya2023expertqa, sun2023vtg, li2023vtgknowledge, kamalloo2023hagrid} simply prompts LLMs (e.g., ChatGPT) to include citations in their generation. \citet{gao2023enabling} conduct systematic evaluations to show that the model performance can often be far from satisfactory with such prompting approaches. 
In this paper, we aim to study how we can \textit{train} an LM to perform better at attributable generation.

One challenge in achieving this goal comes from the expensive cost of collecting supervised training data that contains gold generations interleaved with citations. Another challenge stems from the fine-grained task objective. Instead of a holistic goal of generating the correct response to the task input, attributable generation usually targets on two critical aspects: citation quality and the information correctness of the model response. The citation quality can be further assessed by two dimensions. The first dimension checks if a generated sentence is related to and can be supported by the corresponding cited reference(s). Another dimension inspects if any cited reference in the generation is redundant. 

To address these challenges, we resort to two training algorithms, namely rejection sampling (RS) and reinforcement learning (RL), and propose to use fine-grained rewards as training signals to teach an LM (e.g., LLaMA 2 \citep{touvron2023llama}) to generate attributable responses. We leverage the fine-grained automatic evaluation functions from \citet{gao2023enabling} to provide localized (sentence-level) and specialized (citation quality or response correctness) reward signals as fine-grained rewards. For training, we adapt from the fine-grained RL algorithm in \citet{wu2023fine} and the tree-decoding mechanism in \citet{asai2023selfrag} to perform reinforcement learning and rejection sampling, respectively, using the fine-grained rewards. To ensure a relatively good starting point before RS or RL training, we initialize the LM with distillation from ChatGPT on a handful of sampled training prompts from ALCE \cite{gao2023enabling}.


When evaluating our trained LMs on the test set of ALCE and an out-of-domain dataset EXPERTQA \cite{malaviya2023expertqa}, our main findings are as follows:

\begin{itemize}
    \item Training an LM with fine-grained rewards for attributable generation is much more effective than a holistic reward. This conclusion holds for both RS and RL. To the best of our knowledge, this is the first work that proposes to \textit{train} an LM with \textit{fine-grained} rewards for attributable generation. 
    \item Training with rejection sampling is more efficient and effective than reinforcement learning, while combining them leads to the best model performance that significantly surpasses ChatGPT.
    \item We also show the generalizability of RS and RL that they can produce attributable answers on EXPERTQA, a separate dataset that requires intensive domain knowledge.
\end{itemize}

\section{Problem Definition and Methods}
We follow \citet{gao2023enabling} to define  \textbf{attributable text generation} as: Given a query $x$ and a passage corpus $\mathcal{D}$, generate a textual output $y$ that consists of $N$ sentences $s_1, s_2, \dots, s_N$ 
and citations (such as [1], [1][2], etc.) that refer to passages $\mathcal{C}_i \subseteq \mathcal{D}$ to provide evidence for each sentence $s_i$. 

In this section, we introduce how to train an LM with fine-grained rewards to perform the task. 
As the corpus $\mathcal{D}$ usually contains millions of passages, we first retrieve a small set (e.g., top-5) of relevant passages as the first step before training an LM to generate the output, conditioning on the query and retrieved passages. The details of the retriever for each dataset are in Appendix~\ref{app:retrieval}.
Figure~\ref{fig:method_diagram} illustrates our training process.
As supervision for in-text citations is hard to obtain, we first initialize an LM with distillation from a proprietary LM such as ChatGPT, as described in \S~\ref{sec:distill}. We then introduce how fine-grained rewards are computed and used to train the LM with RL or RS in \S~\ref{sec:fg_train}. Finally, we describe how to train the LM with holistic rewards for comparison purposes in \S~\ref{sec:hol_train}.

\begin{figure*}[ht!]
    \includegraphics[width=\textwidth]{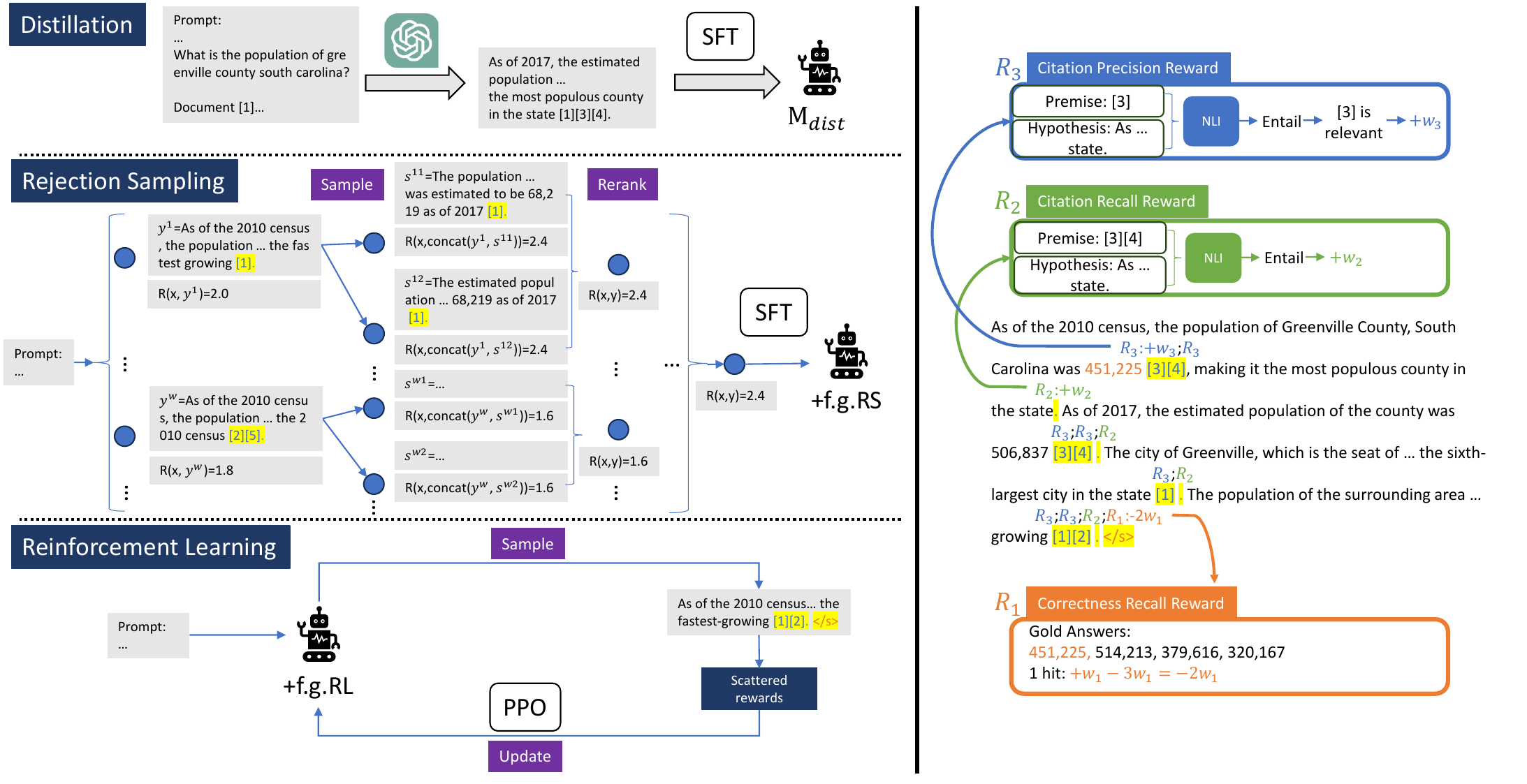}
    \caption{\label{fig:method_diagram} \textbf{Right}: The assignment of our fine-grained rewards (\textcolor{Orange}{$R_1$}: Answer Correctness, \textcolor{OliveGreen}{$R_2$}: Citation Recall, \textcolor{blue}{$R_3$}: Citation Precision). These rewards are assigned to corresponding tokens in the response (citation, EOS Token, etc.; highlighted in \colorbox{yellow}{yellow}). \textbf{Left}: An overview of our framework. Top: Distillation from ChatGPT (\S~\ref{sec:distill}); Middle: Rejection Sampling (\S~\ref{sec:rejection_sampling}); Bottom: Reinforcement Learning (\S~\ref{sec:reinforcement_learning}). 
    }
\end{figure*}

\subsection{Distillation from ChatGPT}
\label{sec:distill}
\citet{gao2023enabling} observe that the performance of open-sourced LMs like LLaMA-2 for attributable generation is far from satisfactory. In particular, we notice that even with in-context learning, a base LLaMA-2-7B can only generate citations for a very small portion of sentences in its responses. As previous studies \citep{touvron2023llama, wu2023fine} often perform RS or RL training on top of a reasonably good initial LM as a starting point, 
we first learn a better LM capable of generating in-text citations by distilling knowledge from a powerful proprietary model such as ChatGPT\footnote{We use gpt-3.5-turbo-0301}. For each question in the training set, we prompt ChatGPT with in-context demonstrations \cite{brown2020language} to generate answers that contain citations (See Appendix~\ref{app:prompts}). These generated answers serve as weak supervisions to finetune our LM. 
We use LLaMA-2-7B as our LM and construct its input as the concatenation of the task instruction, the query $x$, and the passages retrieved for $x$. We denote the resulting model after distillation as $\mathcal{M}_{dist}$.

\subsection{Training with Fine-grained Rewards}
\label{sec:fg_train}

Attributable text generation aims to generate responses that contain correct information and high-quality citations. Its task objective is often decomposed into three aspects \cite{gao2023enabling}: information correctness, citation recall, and citation precision. As a result, we use three reward functions to represent these three sub-goals respectively. Our \textbf{fine-grained rewards} are detailed below: 

\paragraph{Correctness \textcolor{orange}{$R_1$}.} 
Correctness of a model-generated response to a given query is usually measured as the fraction of key information pieces included in the response \cite{stelmakh2022asqa}. Specifically, in most existing factoid QA datasets whose questions require synthesizing multiple documents to answer \cite{stelmakh2022asqa, rubin2022qampari, sun2022conditionalqa}, each question comes with a list of short phrases or important information statements that cover different aspects of the complete answer, referred to as key information list. We can either apply an exact string match (EM) or use a Natural Language Inference (NLI) model to infer whether each item (sub-claim) in the list is covered in the model response. See more in Appendix~\ref{app:correctness_recall}. Note that although in-text citations can be hard to obtain for training, the key information lists are often provided in existing datasets.
For each item in the list, we assign a reward of $+w_1$ if it is in the response, and $-w_1$ otherwise.  
We then sum up the rewards for all list items to obtain the reward for the whole response.

\paragraph{Citation Recall \textcolor{OliveGreen}{$R_2$}.} Citation Recall is the percentage of sentences in the model response that can be supported by the corresponding cited passages. We use an NLI model \cite{honovich2022true} to determine whether each sentence is entailed by its reference passages.
Specifically, the cited passage(s)\footnote{If multiple passages are cited, we concatenate them.} are treated as the premise and the answer sentence is treated as the hypothesis for the NLI model. We assign $+w_2$ for each sentence that can be entailed by the premise, and $-w_2$ otherwise. As a result, \textcolor{OliveGreen}{$R_2$} is computed at the sentence level. See more in Appendix~\ref{app:citation_recall}.

\paragraph{Citation Precision \textcolor{blue}{$R_3$}.} Citation Precision refers to the percentage of citations in the response that can help support their sentences \cite{gao2023enabling}. The same NLI model in \textcolor{OliveGreen}{$R_2$} is used to infer whether each cited passage is necessary to entail the sentence. We assign $+w_3$ for each citation that is "relevant" (See Appenidx~\ref{app:citation_precision}), and $-w_3$ otherwise. Therefore, we apply \textcolor{blue}{$R_3$} at the citation level.

\subsubsection{Rejection Sampling (RS)}
\label{sec:rejection_sampling}
RS is a decoding strategy that can be used to generate labels for model fine-tuning. Specifically, the model is prompted with each training example input to generate multiple output samples. Then a \textit{holistic} reward function is used to rank these samples and the top-1 output is used as the label for model fine-tuning. 

In previous studies \cite{touvron2023llama, nakano2021webgpt}, RS has proven to be effective in improving model performance. In this work, we apply a novel \textit{fine-grained} RS training approach to improve the LM's performance on attributable generation after the distillation step in Section~\ref{sec:distill}. 

We adopt a sentence-level beam search decoding algorithm introduced in \citet{asai2023selfrag} to rank model outputs with our fine-grained reward functions. At each sentence step, the LM generates $K$ different continuation candidates for each sequence in the beam. Then we obtain the top-$B$ generation sequences based on a combined reward, where $B$ is the beam size. We compute the combined reward $R$ as the sum of \textcolor{orange}{$R_1$}, \textcolor{OliveGreen}{$R_2$}, \textcolor{blue}{$R_3$} for each generated sequence $y^w = s_{1, \dots, i}$ at a sentence step. More specifically, $R$ is computed as:
\begin{align}
    \scriptstyle
    R(x,y^w) = \sum_{u=1}^{3}\sum_{k=1}^{l_u}R_u^{k}(x,y^w)
\label{eq:rs_reward}
\end{align}
where $x$ is the query that is inputted together with $y^w$ to compute the rewards of $y^w$, $l_u$ is the number of corresponding segments in $y^w$ for $R_u$ (e.g., $l_1=1$ since $R_1$ treats the entire generation as a single sequence, and $l_2=i$ since $R_2$ is computed at the sentence level). $R_u^{k}$ denotes the reward score $R_u$ for the $k^{th}$ segment in $y^w$.

The whole process stops until all sequences in the beam reaches the end of generation or the maximum number of steps $H$ is reached. Then the highest ranked sequence is returned for further LM fine-tuning as in a standard RS process.


\subsubsection{Reinforcement Learning (RL)}
\label{sec:reinforcement_learning}

We also experiment with LM training using RL after the distillation step. Following \citet{wu2023fine}, we optimize the policy LM using proximal policy optimization (PPO) \cite{schulam2017ppo} with fine-grained rewards. We adapt the token-level reward assignment in \citet{wu2023fine} to our own fine-grained reward functions by assigning i) the sequence-level \textcolor{orange}{$R_1$} to the EOS token of each generated response; ii) the \textcolor{OliveGreen}{$R_2$} score for each sentence to the last token of that sentence; iii) the \textcolor{blue}{$R_3$} score for each bracketed citation to its end token (i.e., the right bracket). If a token is assigned by multiple reward functions, we take their sum to be the final reward for that token.

Formally, given a (prompt, response) pair $(x, y)$, we denote the token time step of the EOS token of $y$ as $T_{1}^{1}$
, the time step of the end token of each sentence $s_i$ as $T_{i}^{2}$, and the time step of the end token for each citation $c_j$ as $T_{j}^{3}$. 
We use $\theta$ and $\theta_{init}$ to represent the current and initial policy respectively and refer to $P$ as the probability distribution over the next token given the generation prefix $g_t$ at time step $t$. The reward for each token $a_t$ generated at time step $t$ in $y$ is:
\begin{align}
    \scriptstyle
    r_t = \sum_{u=1}^{3}
    \sum_{k=1}^{l_u}\left(\mathds{1}(t=T_{k}^{u})R_u^{k}(x,y)\right) - \beta \log\frac{P_\theta(a_t|g_t)}{P_{\theta_{init}}(a_t|g_t)}
\label{eq:rl_reward}
\end{align}
where $l_1$, $l_2$, and $l_3$ denote the number of sequences (which is 1), sentences, and citations in $y$ and $\beta$ is the coefficient for the KL penalty commonly used in PPO. 
$R^k_u$ denotes $R_u$ for the $k^{th}$ segment in $y$.

With the above reward assignment, we follow the standard PPO algorithm to optimize both the policy and value models.

\subsubsection{Combining RS and RL}
\label{sec:combine}

Stacking RL on top of RS is shown to be effective \cite{touvron2023llama} for LLM training. Therefore, we also experiment with this option. 
Starting from $\mathcal{M}_{dist}$, we apply RS to fine-tune the model, and then apply RL to further train it.

\subsection{Training with Holistic Rewards}
\label{sec:hol_train}

For comparison purposes, we also analyze model performances when trained with $holistic$ rewards. In this setting, each generated response is assigned a single overall reward. Similarly, we experiment with both RS and RL for model fine-tuning. We use $R$ in Eq.~\ref{eq:rs_reward} when $i=N$ as our holistic reward.

\paragraph{RS} We follow the standard RS pipeline to have the LM generate multiple outputs and rank them by the holistic reward $R$. The best generated output is then used for model fine-tuning. In contrast, our \textit{fine-grained} RS samples and ranks the generated sequences at each sentence step, which allows finer and better control over the sampling process.

\paragraph{RL} We assign the holistic reward to the last token of each generated sequence and all other tokens receive a reward of 0. In contrast, \textit{fine-grained} RL assigns denser rewards to tokens in a sequence. Similar to Eq.~\ref{eq:rl_reward}, we add the KL penalty term to each token's reward. This RL process is designed in the same way as in most recent work in RL from human preference feedback \cite{menick2022teaching, nakano2021webgpt, touvron2023llama}.
\section{Experiment Setup}
\label{sec:setup}

\subsection{Datasets}
\label{sec:datasets}
We fine-tune our LM with three datasets in the ALCE \cite{gao2023enabling} benchmark designed for the task of attributable generation: (1) ASQA \cite{stelmakh2022asqa}, a long-form QA dataset containing ambiguous questions that can have different interpretations and corresponding answers under different contexts; (2) QAMPARI \cite{rubin2022qampari}, a factoid QA dataset where the answer to each question is a list of entries; (3) ELI5 \cite{fan2019eli5}, a long-form QA dataset with mostly Why/How/What type of questions. Most questions in these datasets require the model to synthesize knowledge from multiple passages to construct a complete answer. 
Each dataset has its own passage corpus, and 
we use GTR \cite{ni2022gtr} or BM25 to retrieve the top-5 passages from the corresponding corpus for each question, following \citet{gao2023enabling}. See more in Appendix~\ref{app:retrieval}.

We train the LM on all of the three QA datasets. We pre-process the training examples to address the imbalance of the three datasets' training sets. This results in roughly 3k/1k train/dev examples evenly distributed over the three datasets. Furthermore, as ALCE only provides development set examples, we obtain around 3k test set examples with roughly 1k instances from each of the ASQA, QAMPARI, and ELI5 test sets.

To verify the generalizability of our framework, we test our model using a dataset unseen during training, EXPERTQA \cite{malaviya2023expertqa}, which is another long-form QA dataset whose questions requires knowledge in specific domains (e.g., psychology) to answer.
See more in Appendix~\ref{app:datasets}.

\subsection{Evaluation Metrics}
\label{sec:metrics}

We use the evaluation metrics in the original ALCE benchmark\footnote{While ALCE also uses MAUVE \cite{pillutla2021mauve} to evaluate fluency, \citet{gao2023enabling} show it is not always reliable. We do not include it in the main analysis.}. \textbf{Correctness Recall} measures the percentage of gold answers (key information pieces) captured by the response, and its implementations vary among datasets (\textbf{EM Rec};\textbf{Rec.-5};\textbf{Claim Rec}). \textbf{Citation Recall} measures the percentage of sentences in the response that are entailed by their cited passages. \textbf{Citation Precision} measures the percentage of citations that can help support the output sentences. On QAMPARI, there is an additional  \textbf{Correctness Precision} metric, which measures the percentage of predicted answer items that are correct, following \cite{gao2023enabling}. See Appendix~\ref{app:metrics} and Figure~\ref{fig:metric_diagram} for more details.

On EXPERTQA, we use the originally designed metrics as well, including \textbf{AutoAIS} \cite{gao2023rarr}, which is similar to Citation Recall and measures the percentage of sentences supported by the citations
, and \textbf{FActscore} \cite{min2023factscore}, which measures the percentage of generated claims that are factual.


\subsection{Training Details}
We fine-tune LLaMA-2-7B \cite{touvron2023llama} in the main experiments. For RS with fine-grained rewards, we use a beam width ($B$) of 8, and we sample 2 continuations ($K$) for each sequence in the beam in a single search step. We set a sentence-level maximum search depth ($H$) of 5 on ASQA and ELI5, and 10 on QAMPARI, since QAMPARI questions have more key information pieces to answer on average. For holistic RS, we set the number of sampling sequences to 16 to make a fair comparison. In terms of reward weights, we set each of $w_1$, $w_2$, and $w_3$ to 0.2 throughout the RS and RL process for LLaMA-2-7B. Empirically, we find that this weight combination serves our goal of \textbf{optimizing citation quality, while ensuring correctness}. See more details in Appendix~\ref{app:training}.

\subsection{Baselines}
As mentioned in \S~\ref{sec:distill}, we initialize the language model (LLaMA-2-7B) with $\mathcal{M}_{dist}$ using the training examples generated from ChatGPT. After this step, we leverage fine-grained rewards to fine-tune the model with rejection sampling (\textbf{f.g. RS}) or reinforcement learning (\textbf{f.g. RL}). We also experiment with combining RS and RL for model fine-tuning (\textbf{f.g. RS+RL}). As discussed in \S~\ref{sec:hol_train}, we also report results of using holistic rewards for RS (\textbf{h. RS}), RL (\textbf{h. RL}), and RS+RL (\textbf{h. RS+RL}). 

We compare the above \textit{fine-tuned} LMs with results from in-context learning (\textbf{ICL}) \cite{brown2020language}. Specifically, we prompt the LM with 2 in-context demonstrations where each contains the question, 5 retrieved passages, and an answer with citations (See Appendix~\ref{app:prompts}). We report ICL results with \textbf{ChatGPT} (GPT-3.5-turbo) or \textbf{LLaMA-2-7B}. 
\section{Results and Analysis}
\label{sec:result}

We first compare all the systems (\S~\ref{sec:main_results}), and then conduct 
ablation study on the training effect of each reward type (\S~\ref{sec:ablation_study}). We also analyze the effect of retrieval on the final model performance (\S~\ref{sec:retrieval_analysis}), and the citation errors that we observe (\S~\ref{sec:citation_error_analysis}). Lastly, we test our models' generalization capability on EXPERTQA that is unseen during training (\S~\ref{sec:generalizability}).

\subsection{Main Results}
\label{sec:main_results}

\begin{table*}[!htp]
	\centering
    \small
	\resizebox{0.8\textwidth}{!}{
    \begin{tabular}{lccc|cccc|ccc}
		\toprule
        & \mc{3}{c}{ASQA} & \mc{4}{c}{QAMPARI} & \mc{3}{c}{ELI5} \\
		& \mc{1}{c}{Correctness} & \mc{2}{c}{Citation} & \mc{2}{c}{Correctness} & \mc{2}{c}{Citation} & \mc{1}{c}{Correctness} & \mc{2}{c}{Citation}\\
		\cmidrule{2-4} \cmidrule{5-8} \cmidrule{9-11}
	  & EM Rec & Rec & Prec & Rec.-5 & Prec & Rec & Prec & Claim Rec & Rec & Prec \\
		\midrule
        ICL {\scriptsize ChatGPT} & 39.96 & 74.72 & 70.97 & 18.34 & 18.57 & 17.65 & \textbf{20.60} & \textbf{13.47} &50.94 &47.58\\
        ICL {\scriptsize LLaMA-2-7B} & 34.15 & 14.12 & 15.26 &8.24&9.23&7.51&7.65&7.83&14.44&8.92\\
        \midrule
        $\mathcal{M}_{dist}$ & 35.56 & 74.80 & 67.99 &17.26&16.18&18.69&18.94&12.03&49.69&45.71\\
        \rowcolor{lightgray}\mc{11}{c}{\textit{RL/RS only}}\\
        ~~~+h.RL & 34.33 & 75.77 & 70.12 & 17.30 & 16.44 & 16.39 & 18.92 & 11.52 & 51.77 & 49.32\\
        ~~~+f.g.RL & 35.99 & 76.30 & 72.38 &\textbf{18.39}&18.81&17.82&19.07&11.60&51.29&51.09\\
        ~~~+h.RS & 37.96 & 74.86 & 68.48 &14.62&15.21&16.71&17.98&11.60&54.10&48.95\\
        ~~~+f.g.RS & \textbf{40.07} & 76.71 & 74.35 & 16.14&18.95&18.56&19.50&11.67&58.75&55.03\\
        \rowcolor{lightgray}\mc{11}{c}{\textit{RS + RL}}\\
        ~~~+h.(RS+RL) & 37.33 & 74.86 & 69.37 & 15.02&15.67&16.82&18.09&11.21&55.62&50.58\\
        ~~~+f.g.(RS+RL) & 40.05 & \textbf{77.83} & \textbf{76.33} & 16.65 &\textbf{19.54}&\textbf{19.50}&20.03&11.54&\textbf{60.86}&\textbf{60.23}\\
		\bottomrule
	\end{tabular}}
	\caption{\label{tab:main} Main result on the test set of three QA datasets. 
    }
\end{table*}

The main results are summarized in Table~\ref{tab:main}.

\paragraph{Training with fine-grained rewards greatly boosts performance.} Adding f.g. RL training to the initial model ($\mathcal{M}_{dist}$), we observe the performance increase on every metric and dataset, except for Citation Recall on QAMPARI and Claim Recall on ELI5. 
Compared to RL, training with RS generally leads to more performance improvement across the three datasets. However, we observe that combining RS and RL leads to the best performance.

In addition, there is a much larger performance boost on ASQA and ELI5 with fine-tuning than on QAMPARI. We suspect the reason behind this to be the fact that gold responses of QAMPARI are lists of entities, while the other two datasets have natural language responses, making training biased towards the latter format. Appendix~\ref{app:main_complete} shows that in the $separate$ training setting (i.e., training LMs on each $individual$ dataset), fine-tuning with either RS, RL or RS+RL can give a significant performance boost for QAMPARI.

Across different f.g. models, we observe a consistent increase in citation recall and precision, while correctness at least remains on the same level if not increase, which matches our objective.

\paragraph{Fine-grained rewards help smaller LLMs surpass ChatGPT.} 
Although the performance gap remains between LLaMA-2 and ChatGPT after distillation, LLaMA-2 trained with fine-grained rewards can close the gap or even outperform ChatGPT. Training with a combined RS and RL leads to an average of 4.0\%, 0.9\%, 10.6\% 
improvements over ChatGPT among all evaluation metrics on ASQA, QAMPARI, and ELI5 respectively. 

\paragraph{Fine-grained rewards are better than holistic rewards.} We compare fine-grained rewards with holistic ones for model training. Table~\ref{tab:main} shows the outstanding effectiveness of fine-grained rewards in all training setups (RS, RL, RS+RL). We observe higher performance gains with fine-grained rewards in almost all metrics and all datasets. The only exception is that the citation recall on ELI5 is similar for h. RL and f.g. RL. The performance gap between using fine-grained and holistic rewards is wider for RS compared to RL.

\subsection{Ablation of Reward Models}
\label{sec:ablation_study}

\begin{table*}[!htp]
	\centering
    \small
    \resizebox{1.0\textwidth}{!}{
	\begin{tabular}{lcccc|ccccc|cccc}
		\toprule
        & \mc{4}{c}{ASQA} & \mc{5}{c}{QAMPARI} & \mc{4}{c}{ELI5} \\
		& \mc{1}{c}{Correctness} & \mc{2}{c}{Citation} & \mc{1}{c}{Length} & \mc{2}{c}{Correctness} & \mc{2}{c}{Citation} & \mc{1}{c}{Length} & \mc{1}{c}{Correctness} & \mc{2}{c}{Citation} & \mc{1}{c}{Length}\\
		\cmidrule{2-5} \cmidrule{6-10} \cmidrule{11-14}
	  & EM Rec & Rec & Prec & \#Tokens & Rec.-5 & Prec & Rec & Prec & \#Tokens & Claim Rec & Rec & Prec & \#Tokens\\
        \midrule
        ICL {\scriptsize ChatGPT} & 39.96 & 74.72 & 70.97 & 37.51 & 18.34 & 18.57 & 17.65 & 20.60 & 13.70 & 13.47 &50.94 &47.58 &86.06 \\
        \midrule
		~~~+f.g.RS$_{sep}$ & 40.24 & 77.65 & 74.96 & 51.34 &17.48&\textbf{20.67}&\textbf{20.62}&\textbf{21.65}&11.24&11.87&61.27&56.45&83.01\\
		~~~w/o. \textcolor{OliveGreen}{$R_2$}+\textcolor{blue}{$R_3$} & \textbf{41.29} & 49.51 & 67.54 & 56.34 & \textbf{23.12} & 18.13 & 15.99 & 16.73 & 15.29 & \textbf{13.60} & 41.06& 43.38 & 88.38\\
		~~~w/o. \textcolor{orange}{$R_1$} & 39.79 & \textbf{79.42} & \textbf{75.69} & 55.89 &12.62&17.63&20.31&21.45&10.91&11.17&\textbf{62.92}&\textbf{58.51}&84.26\\
		\bottomrule
	\end{tabular}}
	\caption{\label{tab:ablation_result} Ablation study results. Our models here are trained on each \textit{individual} dataset.}
	\vspace{-0.2cm}
\end{table*}


We conduct our ablation study to understand how learning to optimize one single task objective (answer correctness or attribution quality) affects the model performance. Since we show in \S~\ref{sec:main_results} that RS generally gives better results than RL, we conduct this ablation analysis on f.g. RS. We train the LM on each \textit{individual} QA dataset by using either the rewards associated with answer correctness (\textcolor{orange}{$R_1$}) or citation quality (\textcolor{OliveGreen}{$R_2$} and \textcolor{blue}{$R_3$}). See results in Table~\ref{tab:ablation_result}.

To show the effect of only optimizing towards answer correctness, we remove \textcolor{OliveGreen}{$R_2$} and \textcolor{blue}{$R_3$} during RS. As indicated by the number of tokens (\#Tokens) metric in Table~\ref{tab:ablation_result}, if the model is only given the correctness reward, it will increase its response length in an attempt to capture more gold answers. Citation quality is ignored and decreases quickly. As a result, the model achieves the highest correctness recall compared to other systems.

We also attempt to focus only on citation quality and remove \textcolor{orange}{$R_1$} during RS. The models produce better results on citation quality on the long-form QA datasets, while their correctness recall drops.

\subsection{Retrieval Analysis}
\label{sec:retrieval_analysis}

Since the retrieved passages may not cover all gold answers \cite{gao2023enabling}, we calibrate the model's correctness metrics, so that we can directly inspect the model's ability to capture answers in the given passages. Specifically, we ignore gold answers that are not recalled by the top 5 retrieved passages when computing correctness metrics. 
We observe that our fine-grained methods (f.g. RL; f.g. RS; f.g.(RS+RL)) can capture around 70\% of whatever is already captured by the passages on ASQA and QAMPARI, and this number becomes more than 90\% on ELI5. In other words, \textbf{our trained LLMs are effective in locating answers within retrieved passages}, although there is still space to improve the models' correctness recall by capturing the remaining answers in the passages. 
On the other hand, to further improve answer correctness, it is important to improve the retrieval performance. 
It might also be possible that a correctness recall gets high when the retrieved passages don't contain gold answers, as is in the case of ELI5 (the Passage Recall is only 9.13, while ChatGPT's Correctness Recall is 13.47), since LLMs have parametric knowledge embedded during pretraining. However, such generated answers are not attributable.
We include more results in Appendix~\ref{app:retrieval_analysis}.


\subsection{Citation Error Analysis}
\label{sec:citation_error_analysis}
We randomly select and inspect responses generated by f.g. RS+RL for 30 test examples (10 from ASQA/QAMPARI/ELI5 each), and observe the following citation errors. 
We include examples for each of the error categories in Appendix~\ref{app:error_analysis_examples}.

\paragraph{Mixing up of passage ids (5.26\%).} This refers to the case where the claim is \textit{unsupported} as a result of the model mixing up the passage ids. The model sometimes \textit{correctly} extracts and paraphrases the content from one passage, but attributes them to another passage. 

\paragraph{Redundant citations (31.58\%).} This refers to the case where passages that are \textit{irrelevant} to the response are cited. An example case is when no retrieved passage is relevant to the question. In this case, the model may generate the response based on its own parametric knowledge, but still cite irrelevant passages. This error is common on the long-form QA datasets\footnote{We explicitly ask the model to attach only 1 citation to each item in the response on QAMPARI following \citet{gao2023enabling}. As a result, this error is less common on QAMPARI.} ASQA and ELI5, and it directly impacts Citation Precision.

\paragraph{Misinterpretation of cited passages (63.16\%).} We notice that our LLMs's responses are more \textit{extractive} after training. However, the model may still interpret and paraphrase the facts mentioned in the passages \textit{incorrectly}. There are two subcategories. The first is \textbf{misinterpretation of a single passage (52.63\%)}, where facts in a single passage are distorted. This error is common on QAMPARI, where the questions usually involve multi-hop relations between multiple entities and the model can easily make mistakes in the reasoning chain. The second is \textbf{incorrect synthesis of multiple passages (10.53\%)}, where the model makes logical errors (e.g., hallucinating relations between entities) when connecting information from different passages.



\subsection{Analysis on Generalizability}
\label{sec:generalizability}

\begin{table}[!htp]
	\centering
    \small
	\setlength{\tabcolsep}{2pt}
	\begin{tabular}{lccc}
		\toprule
        & \mc{3}{c}{EXPERTQA} \\
		\cmidrule{2-2} \cmidrule{3-3} \cmidrule{4-4}
	  & AutoAIS & FActscore & \#Sents\\
		\midrule
        ICL {\scriptsize ChatGPT} & 56.98 & \textbf{85.83} & 8145\\
        ICL {\scriptsize LLaMA-2-7B} & 19.63 & 82.77 & 10053\\
        \midrule
        $\mathcal{M}_{dist}$ & 51.33 & 83.46 & 7293\\
        \rowcolor{lightgray}\mc{4}{c}{\textit{RL/RS only}}\\
        ~~~+h.RL & 53.64 & 83.82 & 8139\\
        ~~~+f.g.RL & 56.15 & 83.89 & 7322\\
        ~~~+h.RS & 57.88 & 83.47 & 8436\\
        ~~~+f.g.RS & 63.49 & 83.85 & 7450\\
        \rowcolor{lightgray}\mc{4}{c}{\textit{RS+RL}}\\
        ~~~+h.(RS+RL) & 59.42 & 83.71 & 8012\\
        ~~~+f.g.(RS+RL) & \textbf{66.12} & 83.78 & 6256\\
		\bottomrule
	\end{tabular}
	\caption{\label{tab:generalizability_result} Result on EXPERTQA. \#Sents: Total number of sentences in the responses to the 2,169 samples.}
	\vspace{-0.2cm}
\end{table}

To demonstrate the generalizability of our model, we evaluate it on a separate dataset, EXPERTQA \cite{malaviya2023expertqa}. We adopt the metrics from its paper, which measure citation recall and factuality, and the results are shown in Table~\ref{tab:generalizability_result}.

\paragraph{The capability of attributable generation is generalizable.} Similar to the previous three datasets, the model trained with fine-grained rewards gives the highest AutoAIS scores \cite{gao2023rarr}, signifying the highest percentage of sentences that are supported by citations. This indicates the robustness and transferability of our model's attribution capability when trained with fine-grained rewards.

\paragraph{Trained LLMs have high factual precision.} While we do not use factuality as a reward model, the FActscore \cite{min2023factscore} metric shows that more than 80\% of sentences generated by our methods are factual, which is close to that of ChatGPT.

\section{Related Work}

\textbf{Retrieval-Augmented Language Models (RALM).}
There exists prior work on RALMs for tasks such as question-answering and fact-checking. 
Aside from the works on kNN-LM \cite{khandelwal2020generalization, zhong2022training}, most of them involve (1) document selection and (2) document reading. Many studies 
attempt to improve document selection by training a stronger retriever or applying additional filtering mechanisms. Methods such as REALM \cite{guu2020retrieval}, RAG \cite{lewis2020rag}, and Atlas \cite{izacard2022atlas} train the retrievers and LMs jointly in an end-to-end setting. \citet{lin2023radit} finetune the LMs and retrievers in two separate stages that complement each other. Others made efforts to train an extra NLI model \cite{yoran2023making} or only the LLMs themselves \cite{luo2023sail, asai2023selfrag} to filter the retrieved documents. Other lines of work explore effective retrieve-on-the-fly methods \cite{parisi2022talm, schick2023toolformer, jiang2023active, ram2023incontext, gao2023enabling} and post-hoc retrieval for answer attribution or revision \cite{gao2023enabling, malaviya2023expertqa, he2022rethinking}.


Most previous work does not include explicit citations or quotes in their models' responses. However, generating citations can both help reduce hallucination and increase verifiability \cite{huang2023citation, li2023llmattribution}. 
Among those prior methods that do provide citations, they either require intensive human annotations such as demonstrations and preferences \cite{nakano2021webgpt} 
, only consider simple questions requiring a single source \cite{menick2022teaching, bohnet2023attributed}, do not achieve satisfactory results on smaller LLMs compared with ChatGPT \cite{gao2023enabling, malaviya2023expertqa, asai2023selfrag}, or use sophisticated mechanism that involves repeated prompting of LLMs and revision of their responses during inference \cite{sun2023vtg, ye2023effective}.
The performance of existing commercial systems like BingChat is neither satisfactory in terms of their citation recall \cite{malaviya2023expertqa, liu2023evaluating}.





\paragraph{Training with Reward Models.}
Recent work uses reward models to train LLMs for various tasks. In the context of RALMs, \citet{menick2022teaching} trains a preference-based reward model and then uses Reinforcement Learning from Human Preference Feedback (RLHF) to train its model to quote knowledge sources. \citet{nakano2021webgpt} train similar reward models, but they also use them for best-of-n sampling during inference on top of RLHF. \citet{zhou2023language} use tree search for QA tasks and utilize an LM value function to generate scores for each solution trajectory. A common characteristic of these reward models is that they only measure the overall quality of a response. A more natural and informative reward model for citations is one that is fine-grained since citations can be scattered throughout a piece of text. Inspired by \citet{wu2023fine}, we divide a holistic reward into more specific ones using both reward density and category.

\section{Conclusion and Future Directions}

In this work, we systematically analyze common LLM training methods for attributable text generation. We show that training with fine-grained rewards gives the strongest performance boost and it also enables small LLMs to outperform ChatGPT.

We propose a number of directions to explore. Firstly, our retrieval analysis shows room to further improve LLMs' correctness recall on QA datasets, and more work can be done to enhance LLMs' reading comprehension and synthesis ability given the retrieved passages. Secondly, according to \citet{touvron2023llama}, iteratively applying imitation learning and reinforcement learning for multiple rounds can continuously increase the model's performance. While this is not the main focus of our work, it will be intriguing to see how far a smaller LLM can go in our task setting using this approach.

\section*{Limitations}
There are several limitations of our work. Firstly, our correctness reward model on ELI5 may not reflect the true correctness recall fully. The sub-claims inferred by text-davinci-003 are shown to be mostly correct, where more than 90 percent of them are faithful to the original long-form answers from which they are derived \cite{gao2023enabling}. However, these sub-claims can be incomplete, meaning that they may not cover every aspect of their sources. This can cause inconsistencies between the training objectives, evaluation metrics, and actual correctness. Inspection of the sub-claims completeness requires intensive human annotations and is deferred to future work.
Secondly, our method still requires an initial distillation step with ChatGPT. This can possibly hinder accessibility when larger, more capable LLMs are not available. A possible resolution to this is to iteratively use In-Context Learning and beam search sampling to bootstrap responses of high quality, and then apply behavioral cloning and reinforcement learning to update the model weights, which can be a future direction to explore.

\section*{Ethics Statement}
The original datasets from which we curate our own train/dev/test splits are open-sourced and publicly available, and thus the risk of sensitive information leakage is kept at a minimum. However, we do acknowledge that our proposed method, especially the corpus that we use for passage retrieval, may introduce certain noises and biases into LLMs' responses. For instance, the corpus that we use on ELI5 and EXPERTQA, Sphere, contains many noisy and biased contents due to its sheer size, and can potentially harm the LLMs' generation. While this is not the main focus of our work, investigating noise-free and socially responsible retrieval mechanisms for attributable text generation is an important future direction.

\bibliography{anthology,custom}

\begin{thebibliography}{48}
\expandafter\ifx\csname natexlab\endcsname\relax\def\natexlab#1{#1}\fi

\bibitem[{Asai et~al.(2023)Asai, Wu, Wang, Sil, and Hajishirzi}]{asai2023selfrag}
Akari Asai, Zeqiu Wu, Yizhong Wang, Avirup Sil, and Hannaneh Hajishirzi. 2023.
\newblock \href {https://arxiv.org/abs/2310.11511} {Self-rag: Learning to retrieve, generate, and critique through self-reflection}.
\newblock \emph{arXiv preprint arXiv:2310.11511, 2023}.

\bibitem[{Bohnet et~al.(2023)Bohnet, Tran, Verga, Aharoni, Andor, Soares, Ciaramita, Eisenstein, Ganchev, Herzig, Hui, Kwiatkowski, Ma, Ni, Saralegui, Schuster, Cohen, Collins, Das, Metzler, Petrov, and Webster}]{bohnet2023attributed}
Bernd Bohnet, Vinh~Q. Tran, Pat Verga, Roee Aharoni, Daniel Andor, Livio~Baldini Soares, Massimiliano Ciaramita, Jacob Eisenstein, Kuzman Ganchev, Jonathan Herzig, Kai Hui, Tom Kwiatkowski, Ji~Ma, Jianmo Ni, Lierni~Sestorain Saralegui, Tal Schuster, William~W. Cohen, Michael Collins, Dipanjan Das, Donald Metzler, Slav Petrov, and Kellie Webster. 2023.
\newblock \href {https://arxiv.org/pdf/2212.08037} {Attributed question answering: Evaluation and modeling for attributed large language models}.
\newblock \emph{arXiv preprint arXiv:2212.08037}.

\bibitem[{Brown et~al.(2020)Brown, Mann, Ryder, Subbiah, Kaplan, Dhariwal, Neelakantan, Shyam, Sastry, Askell, Agarwal, Herbert-Voss, Krueger, Henighan, Child, Aditya~Ramesh, Wu, Winter, Hesse, Chen, Sigler, Litwin, Gray, Chess, Clark, Berner, McCandlish, Radford, Sutskever, and Amodei}]{brown2020language}
Tom Brown, Benjamin Mann, Nick Ryder, Melanie Subbiah, Jared~D Kaplan, Prafulla Dhariwal, Arvind Neelakantan, Pranav Shyam, Girish Sastry, Amanda Askell, Sandhini Agarwal, Ariel Herbert-Voss, Gretchen Krueger, Tom Henighan, Rewon Child, Daniel~Ziegler Aditya~Ramesh, Jeffrey Wu, Clemens Winter, Chris Hesse, Mark Chen, Eric Sigler, Mateusz Litwin, Scott Gray, Benjamin Chess, Jack Clark, Christopher Berner, Sam McCandlish, Alec Radford, Ilya Sutskever, and Dario Amodei. 2020.
\newblock \href {https://papers.nips.cc/paper/2020/hash/1457c0d6bfcb4967418bfb8ac142f64a-Abstract.html} {Language models are few-shot learners}.
\newblock In \emph{Advances in Neural Information Processing Systems}.

\bibitem[{Dettmers et~al.(2023)Dettmers, Pagnoni, Holtzman, and Zettlemoyer}]{dettmers2023qlora}
Tim Dettmers, Artidoro Pagnoni, Ari Holtzman, and Luke Zettlemoyer. 2023.
\newblock \href {https://arxiv.org/abs/2305.14314} {Qlora: Efficient finetuning of quantized llms}.
\newblock \emph{arXiv preprint arXiv:2305.14314}.

\bibitem[{Fan et~al.(2019)Fan, Jernite, Perez, Grangier, Weston, and Auli}]{fan2019eli5}
Angela Fan, Yacine Jernite, Ethan Perez, David Grangier, Jason Weston, and Michael Auli. 2019.
\newblock \href {https://aclanthology.org/P19-1346/} {Eli5: Long form question answering}.
\newblock In \emph{Association for Computational Linguistics (ACL)}, pages 3558–--3567.

\bibitem[{Gao et~al.(2023{\natexlab{a}})Gao, Dai, Pasupat, Chen, Chaganty, Fan, Zhao, Lao, Lee, Juan, and Guu}]{gao2023rarr}
Luyu Gao, Zhuyun Dai, Panupong Pasupat, Anthony Chen, Arun~Tejasvi Chaganty, Yicheng Fan, Vincent Zhao, Ni~Lao, Hongrae Lee, Da-Cheng Juan, and Kelvin Guu. 2023{\natexlab{a}}.
\newblock \href {https://aclanthology.org/2023.acl-long.910} {Rarr: Researching and revising what language models say, using language models}.
\newblock In \emph{Proceedings of the 61st Annual Meeting of the Association for Computational Linguistics (Volume 1: Long Papers)}, page 16477–16508.

\bibitem[{Gao et~al.(2023{\natexlab{b}})Gao, Yen, Yu, and Chen}]{gao2023enabling}
Tianyu Gao, Howard Yen, Jiatong Yu, and Danqi Chen. 2023{\natexlab{b}}.
\newblock \href {https://aclanthology.org/2023.emnlp-main.398} {Enabling large language models to generate text with citations}.
\newblock In \emph{Proceedings of the 2023 Conference on Empirical Methods in Natural Language Processing}, pages 6465–--6488.

\bibitem[{Guu et~al.(2020)Guu, Lee, Tung, Pasupat, and Chang}]{guu2020retrieval}
Kelvin Guu, Kenton Lee, Zora Tung, Panupong Pasupat, and Mingwei Chang. 2020.
\newblock \href {https://dl.acm.org/doi/pdf/10.5555/3524938.3525306} {Realm: Retrieval augmented language model pre-training}.
\newblock In \emph{Proceedings of International Conference on Machine Learning, 2020}, pages 33--40.

\bibitem[{He et~al.(2022)He, Zhang, and Roth}]{he2022rethinking}
Hangfeng He, Hongming Zhang, and Dan Roth. 2022.
\newblock \href {https://arxiv.org/abs/2301.00303} {Rethinking with retrieval: Faithful large language model inference}.
\newblock \emph{arXiv preprint arXiv:2301.00303, 2022}.

\bibitem[{Honovich et~al.(2022)Honovich, Aharoni, Herzig, Taitelbaum, Kukliansy, Cohen, Scialom, Szpektor, Hassidim, and Matias}]{honovich2022true}
Or~Honovich, Roee Aharoni, Jonathan Herzig, Hagai Taitelbaum, Doron Kukliansy, Vered Cohen, Thomas Scialom, Idan Szpektor, Avinatan Hassidim, and Yossi Matias. 2022.
\newblock \href {https://aclanthology.org/2022.naacl-main.287/} {True: Re-evaluating factual consistency evaluation}.
\newblock In \emph{North American Chapter of the Association for Computational Linguistics: Human Language Technologies (NAACL-HLT)}, pages 3905--3920.

\bibitem[{Hu et~al.(2021)Hu, Shen, Wallis, Allen-Zhu, Li, Wang, Wang, and Chen}]{hu2021lora}
Edward~J. Hu, Yelong Shen, Phillip Wallis, Zeyuan Allen-Zhu, Yuanzhi Li, Shean Wang, Lu~Wang, and Weizhu Chen. 2021.
\newblock \href {https://arxiv.org/abs/2106.09685} {Lora: Low-rank adaptation of large language models}.
\newblock \emph{arXiv preprint arXiv:2106.09685, 2021}.

\bibitem[{Huang and Chang(2023)}]{huang2023citation}
Jie Huang and Kevin Chen-Chuan Chang. 2023.
\newblock \href {https://arxiv.org/pdf/2307.02185} {Citation: A key to building responsible and accountable large language models}.
\newblock \emph{arXiv preprint arXiv:2307.02185}.

\bibitem[{Izacard et~al.(2022)Izacard, Lewis, Lomeli, Hosseini, Petroni, Schick, Dwivedi-Yu, Joulin, Riedel, and Grave}]{izacard2022atlas}
Gautier Izacard, Patrick Lewis, Maria Lomeli, Lucas Hosseini, Fabio Petroni, Timo Schick, Jane Dwivedi-Yu, Armand Joulin, Sebastian Riedel, and Edouard Grave. 2022.
\newblock \href {https://arxiv.org/abs/2208.03299} {Atlas: Few-shot learning with retrieval augmented language models}.
\newblock \emph{arXiv preprint arXiv:2208.03299, 2022}.

\bibitem[{Ji et~al.(2022)Ji, Lee, Frieske, Yu, Su, Xu, Ishii, Bang, Madotto, and Fung}]{ji2022survey}
Ziwei Ji, Nayeon Lee, Rita Frieske, Tiezheng Yu, Dan Su, Yan Xu, Etsuko Ishii, Ye~Jin Bang, Andrea Madotto, and Pascale Fung. 2022.
\newblock \href {https://dl.acm.org/doi/pdf/10.1145/3571730} {Survey of hallucination in natural language generation}.
\newblock \emph{ACM Computing Surveys}, 55(12):1--–38.

\bibitem[{Jiang et~al.(2023)Jiang, Xu, Gao, Sun, Liu, Dwivedi-Yu, and Yang}]{jiang2023active}
Zhengbao Jiang, Frank~F Xu, Luyu Gao, Zhiqing Sun, Qian Liu, Jane Dwivedi-Yu, and Yiming Yang. 2023.
\newblock \href {https://arxiv.org/abs/2305.06983} {Active retrieval augmented generation}.
\newblock \emph{arXiv preprint arXiv:2305.06983, 2023}.

\bibitem[{Kamalloo et~al.(2023)Kamalloo, Jafari, Zhang, Thakur, and Lin}]{kamalloo2023hagrid}
Ehsan Kamalloo, Aref Jafari, Xinyu Zhang, Nandan Thakur, and Jimmy Lin. 2023.
\newblock \href {https://arxiv.org/pdf/2307.16883} {{HAGRID}: A human-llm collaborative dataset for generative information-seeking with attribution}.
\newblock \emph{arXiv:2307.16883}.

\bibitem[{Khandelwal et~al.(2020)Khandelwal, Levy, Jurafsky, Zettlemoyer, and Lewis}]{khandelwal2020generalization}
Urvashi Khandelwal, Omer Levy, Dan Jurafsky, Luke Zettlemoyer, and Mike Lewis. 2020.
\newblock \href {https://openreview.net/forum?id=HklBjCEKvH} {Generalization through memorization: Nearest neighbor language models}.
\newblock In \emph{International Conference on Learning Representations}.

\bibitem[{Kwon et~al.(2023)Kwon, Li, Zhuang, Sheng, Zheng, Yu, Gonzalez, Zhang, and Stoica}]{kwon2023efficient}
Woosuk Kwon, Zhuohan Li, Siyuan Zhuang, Ying Sheng, Lianmin Zheng, Cody~Hao Yu, Joseph~E. Gonzalez, Hao Zhang, and Ion Stoica. 2023.
\newblock \href {https://dl.acm.org/doi/10.1145/3600006.3613165} {Efficient memory management for large language model serving with pagedattention}.
\newblock In \emph{Proceedings of the ACM SIGOPS 29th Symposium on Operating Systems Principles}.

\bibitem[{Lewis et~al.(2020)Lewis, Perez, Piktus, Petroni, Karpukhin, Goyal, Kuttler, Lewis, tau Yih, Rocktäschel, Riedel, and Kiela}]{lewis2020rag}
Patrick Lewis, Ethan Perez, Aleksandra Piktus, Fabio Petroni, Vladimir Karpukhin, Naman Goyal, Heinrich Kuttler, Mike Lewis, Wen tau Yih, Tim Rocktäschel, Sebastian Riedel, and Douwe Kiela. 2020.
\newblock \href {https://proceedings.neurips.cc/paper/2020/file/6b493230205f780e1bc26945df7481e5-Paper.pdf} {Retrieval-augmented generation for knowledge-intensive nlp tasks}.
\newblock In \emph{Advances in Neural Information Processing Systems}.

\bibitem[{Li et~al.(2023{\natexlab{a}})Li, Sun, Hu, Liu, Chen, Hu, Wu, and Zhang}]{li2023llmattribution}
Dongfang Li, Zetian Sun, Xinshuo Hu, Zhenyu Liu, Ziyang Chen, Baotian Hu, Aiguo Wu, and Min Zhang. 2023{\natexlab{a}}.
\newblock \href {https://arxiv.org/pdf/2311.03731} {A survey of large language models attribution}.
\newblock \emph{arXiv:2311.03731}.

\bibitem[{Li et~al.(2023{\natexlab{b}})Li, Cao, Pan, Ma, and Sun}]{li2023vtgknowledge}
Xinze Li, Yixin Cao, Liangming Pan, Yubo Ma, and Aixin Sun. 2023{\natexlab{b}}.
\newblock \href {https://arxiv.org/abs/2310.05634} {Towards verifiable generation: A benchmark for knowledge-aware language model attribution}.
\newblock \emph{arXiv preprint arXiv:2310.05634}.

\bibitem[{Lin et~al.(2022)Lin, Chen, Chen, Shi, Lomeli, James, Rodriguez, Kahn, Szilvasy, Lewis, Zettlemoyer, and Yih}]{lin2023radit}
Xi~Victoria Lin, Xilun Chen, Mingda Chen, Weijia Shi, Maria Lomeli, Rich James, Pedro Rodriguez, Jacob Kahn, Gergely Szilvasy, Mike Lewis, Luke Zettlemoyer, and Scott Yih. 2022.
\newblock \href {https://arxiv.org/abs/2310.01352} {Ra-dit: Retrievalaugmented dual instruction tuning}.
\newblock \emph{arXiv preprint arXiv:2310.01352, 2022}.

\bibitem[{Liu et~al.(2023)Liu, Zhang, and Liang}]{liu2023evaluating}
Nelson~F. Liu, Tianyi Zhang, and Percy Liang. 2023.
\newblock \href {https://arxiv.org/pdf/2304.09848} {Evaluating verifiability in generative search engines}.
\newblock \emph{arXiv preprint arXiv:2304.09848}.

\bibitem[{Luo et~al.(2023)Luo, Chuang, Gong, Zhang, Kim, Wu, Fox, Meng, and Glass}]{luo2023sail}
Hongyin Luo, Yung-Sung Chuang, Yuan Gong, Tianhua Zhang, Yoon Kim, Xixin Wu, Danny Fox, Helen Meng, and James Glass. 2023.
\newblock \href {https://arxiv.org/abs/2305.15225} {Sail: Search-augmented instruction learning}.
\newblock \emph{arXiv preprint arXiv:2305.15225, 2023}.

\bibitem[{Malaviya et~al.(2023)Malaviya, Lee, Chen, Sieber, Yatskar, and Roth}]{malaviya2023expertqa}
Chaitanya Malaviya, Subin Lee, Sihao Chen, Elizabeth Sieber, Mark Yatskar, and Dan Roth. 2023.
\newblock \href {https://arxiv.org/pdf/2309.07852} {Expertqa: Expert-curated questions and attributed answers}.
\newblock \emph{arXiv preprint arXiv:2309.07852, 2023}.

\bibitem[{Menick et~al.(2022)Menick, Trebacz, Mikulik, Aslanides, Song, Chadwick, Glaese, Young, Campbell-Gillingam, Irving, and McAleese}]{menick2022teaching}
Jacob Menick, Maja Trebacz, Vladimir Mikulik, John Aslanides, Francis Song, Martin Chadwick, Mia Glaese, Susannah Young, Lucy Campbell-Gillingam, Geoffrey Irving, and Nat McAleese. 2022.
\newblock \href {https://arxiv.org/pdf/2203.11147} {Teaching language models to support answers with verified quotes}.
\newblock \emph{arXiv preprint arXiv:2203.11147, 2022}.

\bibitem[{Min et~al.(2023)Min, Krishna, Lyu, Lewis, tau Yih, Koh, Iyyer, Zettlemoyer, and Hajishirzi}]{min2023factscore}
Sewon Min, Kalpesh Krishna, Xinxi Lyu, Mike Lewis, Wen tau Yih, Pang~Wei Koh, Mohit Iyyer, Luke Zettlemoyer, and Hannaneh Hajishirzi. 2023.
\newblock \href {https://arxiv.org/abs/2305.14251v1} {Factscore: Fine-grained atomic evaluation of factual precision in long form text generation}.
\newblock \emph{arXiv preprint arXiv:2305.14251v1}.

\bibitem[{Nakano et~al.(2021)Nakano, Hilton, Balaji, Wu, Ouyang, Kim, Hesse, Jain, Kosaraju, Saunders, Jiang, Cobbe, Eloundou, Krueger, Button, Knight, Chess, and Schulman}]{nakano2021webgpt}
Reiichiro Nakano, Jacob Hilton, Suchir Balaji, Jeff Wu, Long Ouyang, Christina Kim, Christopher Hesse, Shantanu Jain, Vineet Kosaraju, William Saunders, Xu~Jiang, Karl Cobbe, Tyna Eloundou, Gretchen Krueger, Kevin Button, Matthew Knight, Benjamin Chess, and John Schulman. 2021.
\newblock \href {https://arxiv.org/abs/2112.09332} {Webgpt: Browser-assisted question answering with human feedback}.
\newblock \emph{arXiv preprint arXiv:2112.09332, 2021}.

\bibitem[{Ni et~al.(2022)Ni, Qu, Lu, Dai, Abrego, Ma, Zhao, Luan, Hall, Chang, and Yang}]{ni2022gtr}
Jianmo Ni, Chen Qu, Jing Lu, Zhuyun Dai, Gustavo~Hernandez Abrego, Ji~Ma, Vincent Zhao, Yi~Luan, Keith Hall, Ming-Wei Chang, and Yinfei Yang. 2022.
\newblock \href {https://aclanthology.org/2022.emnlp-main.669} {Large dual encoders are generalizable retrievers}.
\newblock In \emph{Proceedings of the 2022 Conference on Empirical Methods in Natural Language Processing}, pages 9844–--9855.

\bibitem[{Parisi et~al.(2022)Parisi, Zhao, and Fiedel}]{parisi2022talm}
Aaron Parisi, Yao Zhao, and Noah Fiedel. 2022.
\newblock \href {https://arxiv.org/abs/2205.12255} {Talm: Tool augmented language models}.
\newblock \emph{arXiv preprint arXiv:2205.12255, 2022}.

\bibitem[{Piktus et~al.(2021)Piktus, Petroni, Karpukhin, Okhonko, Broscheit, Izacard, Lewis, Oğuz, Grave, tau Yih, and et~al.}]{piktus2021sphere}
Aleksandra Piktus, Fabio Petroni, Vladimir Karpukhin, Dmytro Okhonko, Samuel Broscheit, Gautier Izacard, Patrick Lewis, Barlas Oğuz, Edouard Grave, Wen tau Yih, and et~al. 2021.
\newblock \href {https://arxiv.org/abs/2112.09924} {The web is your oyster - knowledge-intensive nlp against a very large web corpus}.
\newblock \emph{arXiv preprint arXiv:2112.09924, 2021}.

\bibitem[{Pillutla et~al.(2021)Pillutla, Swayamdipta, Zellers, Thickstun, Welleck, Choi, and Harchaoui}]{pillutla2021mauve}
Krishna Pillutla, Swabha Swayamdipta, Rowan Zellers, John Thickstun, Sean Welleck, Yejin Choi, and Zaid Harchaoui. 2021.
\newblock \href {https://proceedings.neurips.cc/paper/2021/file/260c2432a0eecc28ce03c10dadc078a4-Paper.pdf} {Mauve: Measuring the gap between neural text and human text using divergence frontiers}.
\newblock In \emph{Advances in Neural Information Processing Systems}.

\bibitem[{Raffel et~al.(2020)Raffel, Shazeer, Roberts, Lee, Narang, Matena, Zhou, Li, and Liu}]{raffel2020t5}
Colin Raffel, Noam Shazeer, Adam Roberts, Katherine Lee, Sharan Narang, Michael Matena, Yanqi Zhou, Wei Li, and Peter~J Liu. 2020.
\newblock \href {https://jmlr.org/papers/v21/20-074.html} {Exploring the limits of transfer learning with a unified text-to-text transformer}.
\newblock \emph{The Journal of Machine Learning Research (JMLR)}, 21(140).

\bibitem[{Ram et~al.(2023)Ram, Levine, Dalmedigos, Muhlgay, Shashua, Leyton-Brown, and Shoham}]{ram2023incontext}
Ori Ram, Yoav Levine, Itay Dalmedigos, Dor Muhlgay, Amnon Shashua, Kevin Leyton-Brown, and Yoav Shoham. 2023.
\newblock \href {https://arxiv.org/abs/2302.00083} {In-context retrieval-augmented language models}.
\newblock In \emph{Transactions of the Association for Computational Linguistics, 2023}.

\bibitem[{Rubin et~al.(2022)Rubin, Yoran, Wolfson, Herzig, and Berant}]{rubin2022qampari}
Samuel Joseph Amouyal~Ohad Rubin, Ori Yoran, Tomer Wolfson, Jonathan Herzig, and Jonathan Berant. 2022.
\newblock \href {https://arxiv.org/abs/2205.12665} {Qampari: An open-domain question answering benchmark for questions with many answers from multiple paragraphs}.
\newblock \emph{arXiv preprint arXiv:2205.12665, 2022}.

\bibitem[{Santacroce et~al.(2023)Santacroce, Lu, Yu, Li, and Shen}]{santacroce2023efficient}
Michael Santacroce, Yadong Lu, Han Yu, Yuanzhi Li, and Yelong Shen. 2023.
\newblock \href {https://arxiv.org/abs/2309.00754} {Efficient rlhf: Reducing the memory usage of ppo}.
\newblock \emph{arXiv preprint arXiv:2309.00754, 2023}.

\bibitem[{Schick et~al.(2023)Schick, Dwivedi-Yu, Dessì, Raileanu, Lomeli, Zettlemoyer, Cancedda, and Scialom}]{schick2023toolformer}
Timo Schick, Jane Dwivedi-Yu, Roberto Dessì, Roberta Raileanu, Maria Lomeli, Luke Zettlemoyer, Nicola Cancedda, and Thomas Scialom. 2023.
\newblock \href {https://arxiv.org/abs/2302.04761} {Toolformer: Language models can teach themselves to use tools}.
\newblock \emph{arXiv preprint arXiv:2302.04761, 2023}.

\bibitem[{Schulman et~al.(2017)Schulman, Wolski, Dhariwal, Radford, and Klimov}]{schulam2017ppo}
John Schulman, Filip Wolski, Prafulla Dhariwal, Alec Radford, and Oleg Klimov. 2017.
\newblock \href {https://arxiv.org/abs/1707.06347} {Proximal policy optimization algorithms}.
\newblock \emph{arXiv preprint arXiv:1707.06347, 2017}.

\bibitem[{Stelmakh et~al.(2022)Stelmakh, Luan, Dhingra, and Chang}]{stelmakh2022asqa}
Ivan Stelmakh, Yi~Luan, Bhuwan Dhingra, and Ming-Wei Chang. 2022.
\newblock \href {https://arxiv.org/abs/2204.06092} {Asqa: Factoid questions meet long-form answers}.
\newblock \emph{arXiv preprint arXiv:2204.06092, 2022}.

\bibitem[{Sun et~al.(2022)Sun, Cohen, and Salakhutdinov}]{sun2022conditionalqa}
Haitian Sun, William Cohen, and Ruslan Salakhutdinov. 2022.
\newblock \href {https://aclanthology.org/2022.acl-long.253} {Conditionalqa: A complex reading comprehension dataset with conditional answers}.
\newblock In \emph{Proceedings of the 60th Annual Meeting of the Association for Computational Linguistics}, page 3627–3637.

\bibitem[{Sun et~al.(2023)Sun, Cai, Wang, Hou, Wei, Wang, Zhang, and Yin}]{sun2023vtg}
Hao Sun, Hengyi Cai, Bo~Wang, Yingyan Hou, Xiaochi Wei, Shuaiqiang Wang, Yan Zhang, and Dawei Yin. 2023.
\newblock \href {https://arxiv.org/abs/2312.09075} {Towards verifiable text generation with evolving memory and self-reflection}.
\newblock \emph{arXiv preprint arXiv:2312.09075}.

\bibitem[{Touvron et~al.(2023)Touvron, Martin, Stone, Albert, Almahairi, Babaei, Bashlykov, Batra, Bhargava, Bhosale, and et~al.}]{touvron2023llama}
Hugo Touvron, Louis Martin, Kevin Stone, Peter Albert, Amjad Almahairi, Yasmine Babaei, Nikolay Bashlykov, Soumya Batra, Prajjwal Bhargava, Shruti Bhosale, and et~al. 2023.
\newblock \href {https://arxiv.org/abs/2307.09288} {Llama 2: Open foundation and fine-tuned chat models}.
\newblock \emph{arXiv preprint arXiv:2307.09288, 2023}.

\bibitem[{Wu et~al.(2023)Wu, Hu, Shi, Dziri, Suhr, Ammanabrolu, Smith, Ostendorf, and Hajishirzi}]{wu2023fine}
Zeqiu Wu, Yushi Hu, Weijia Shi, Nouha Dziri, Alane Suhr, Prithviraj Ammanabrolu, Noah~A Smith, Mari Ostendorf, and Hannaneh Hajishirzi. 2023.
\newblock \href {https://openreview.net/pdf?id=CSbGXyCswu} {Fine-grained human feedback gives better rewards for language model training}.
\newblock In \emph{Advances in Neural Information Processing Systems}.

\bibitem[{Ye et~al.(2023)Ye, Sun, Arik, and Pfister}]{ye2023effective}
Xi~Ye, Ruoxi Sun, Sercan~Ö. Arik, and Tomas Pfister. 2023.
\newblock \href {https://arxiv.org/abs/2311.09533} {Effective large language model adaptation for improved grounding}.
\newblock \emph{arXiv preprint arXiv:2311.09533}.

\bibitem[{Yoran et~al.(2023)Yoran, Wolfson, Ram, and Berant}]{yoran2023making}
Ori Yoran, Tomer Wolfson, Ori Ram, and Jonathan Berant. 2023.
\newblock \href {https://arxiv.org/abs/2310.01558} {Making retrieval-augmented language models robust to irrelevant context}.
\newblock \emph{arXiv preprint arXiv:2310.01558, 2023}.

\bibitem[{Yue et~al.(2023)Yue, Wang, Zhang, Chen, Su, and Sun}]{yue2023automatic}
Xiang Yue, Boshi Wang, Kai Zhang, Ziru Chen, Yu~Su, and Huan Sun. 2023.
\newblock \href {https://arxiv.org/abs/2305.06311} {Automatic evaluation of attribution by large language models}.
\newblock \emph{arXiv preprint arXiv:2305.06311}.

\bibitem[{Zhong et~al.(2022)Zhong, Lei, and Chen}]{zhong2022training}
Zexuan Zhong, Tao Lei, and Danqi Chen. 2022.
\newblock \href {https://aclanthology.org/2022.emnlp-main.382} {Training language models with memory augmentation}.
\newblock In \emph{Proceedings of the 2022 Conference on Empirical Methods in Natural Language Processing}, pages 5657--5673.

\bibitem[{Zhou et~al.(2023)Zhou, Yan, Shlapentokh-Rothman, Wang, and Wang}]{zhou2023language}
Andy Zhou, Kai Yan, Michal Shlapentokh-Rothman, Haohan Wang, and Yu-Xiong Wang. 2023.
\newblock \href {https://arxiv.org/abs/2310.04406} {Language agent tree search unifies reasoning acting and planning in language models}.
\newblock \emph{arXiv preprint arXiv:2310.04406, 2023}.

\end{thebibliography}
\bibliographystyle{acl_natbib}

\appendix

\newpage
\section{Details on Fine-grained Rewards}
\subsection{Correctness Recall Reward}
\label{app:correctness_recall}
In ASQA, each answer (key information piece) corresponds to a separate interpretation of the question and is usually a single phrase. In QAMPARI, each question has a list of gold answers, and each answer is also a single phrase. For these two datasets, we check how many answers from the gold answer set (key information list) are exact substrings of the model response. In ELI5 and EXPERTQA, since only a single long-form answer is available for each question, we infer three sub-claims from the gold long-form answers with text-davinci-003, and use these claims as the key information pieces. We then use an NLI model to check how many sub-claims are entailed by the model response.

In QAMPARI, since users usually may not want the entire list, we do not penalize a missing answer if at least 5 answer entries have been captured. 

As there is no clear boundary for an answer unit in long-form response, we adopt a sequence-level granularity for \textcolor{orange}{$R_1$}, meaning that the reward is computed once and assigned to a single token for the entire response sequence.

Specifically, if the response $y$ to a question from dataset $d$ string matches (ASQA; QAMPARI) or entails (ELI5; EXPERTQA) $h$ items out of $t$ gold answer items, then the reward is
\begin{align}
    \scriptsize
    R_1^1(x,y) = \left\{
        \begin{array}{ll}
            w_1h - w_1(t-h), d\neq qampari\\    
            w_1h - w_1max(min(t, 5)-h, 0), d=qampari\\ 
        \end{array}
    \right.
\end{align}

\subsection{Citation Recall Reward}
\label{app:citation_recall}

Given a sentence $s_i$, its set of cited passages $C_i$, and an NLI model $\sigma$ that checks whether a set of cited documents entail a sentence, $s_i$ has a Citation Recall reward of $+w_2$ (i.e., $R_2^i(x,y)=+w_2$) if and only if it is entailed by the concatenation of the cited passages, meaning that $\sigma(concat(C_i), s_i) = 1$. Otherwise, the reward is $-w_2$ (i.e., $R_2^i(x,y)=-w_2$).

On long-form QA datasets (ASQA; ELI5; EXPERTQA), this reward is computed for each sentence in the response. On QAMPARI, since the model response is a comma-separated list, we treat each comma-separated item (formatted as ’\{question\} \{item\}’) in the response as a sentence.

\subsection{Citation Precision Reward}
\label{app:citation_precision}
We use the definition in \cite{gao2023enabling}. Given a sentence $s_i$, its set of cited passages $C_i$, and an NLI model $\sigma$, a citation that refers to passage $c_j \in C_i$ has a Citation Precision reward of $+w_3$ (i.e., $R_3^j(x,y)=+w_3$) if and only if $\sigma(concat(C_i), s_i) = 1$ and either of the following conditions holds

\begin{itemize}
    \item $c_j$ explicitly entails s with $\sigma(c_j, s) = 1$
    \item $c_j$ implicitly helps other passages entail s with $\sigma(concat(C_i - \{c_j\}), s) = 0$
\end{itemize}

Otherwise, the reward is $-w_3$ (i.e., $R_3^j(x,y)=-w_3$).

\section{Datasets and Metrics}
\label{app:datasets_and_metrics}
\subsection{Datasets}
\label{app:datasets}

\paragraph{ASQA} The authors of ASQA have not released its test split. Therefore, we randomly shuffle and then divide its original train split samples into two groups so that one group is used for training and the other is used for validation. We use its original dev split as our test split. This gives 3,853/500/948 samples for train/dev/test respectively.
\paragraph{QAMPARI} We randomly sample 4,000 instances from the original train split of QAMPARI and take its entire development and test splits, giving 4,000/1,000/1,000 samples for train/dev/test usage.
\paragraph{ELI5} The original ELI5 dataset is much larger than ASQA. To ensure the balance between different datasets, we randomly sample 3,728/1,000/1,000 instances from the train/dev/test splits of ELI5. We also include in the train split 264 samples from WebGPT \cite{nakano2021webgpt}, since these samples' answers contain citations and are of relatively high quality. This gives 3,992/1,000/1,000 samples in the train/dev/test splits.

The above gives around 4k/0.5k/1k samples for train/valid/test splits on ASQA and approximately 4k/1k/1k samples on QAMPARI and ELI5. They are used directly in the \textit{separate} setting (i.e., training LMs on each $individual$ dataset). In the \textit{combined} setting (i.e., training LMs on the mixture; This is our main experiment setting), we further sample around 1k/334/1k instances from each filtered dataset to get 2,992/1,002/3,000 samples for joint training and inference.

\paragraph{EXPERTQA} We remove 8 samples that do not have human-revised answers, which gives us the remaining 2,169 samples. The reason is that we need annotated answers to infer sub-claims for correctness recall computation in the additional experiments in Appendix~\ref{app:expertqa} and we want our testing set in \S~\ref{sec:generalizability} to be consistent with what is used in the additional experiments. Since some groups of samples in EXPERTQA are evaluated in a closed-book setting or use a different retrieval mechanism, we manually retrieve the top 5 passages from Sphere \cite{piktus2021sphere} 
again for each sample to ensure consistency with other datasets. We use EXPERTQA for testing only.

\subsection{Retrieval}
\label{app:retrieval}
We use GTR \cite{ni2022gtr} to retrieve the top-5 passages from the 2018-12-20 Wikipedia snapshot on ASQA and QAMPARI, and use BM25 to retrieve the top-5 passages from Sphere \cite{piktus2021sphere} on ELI5 and EXPERTQA respectively.

\subsection{Metrics}
\label{app:metrics}

\begin{figure*}[ht!]
    \includegraphics[width=\textwidth]{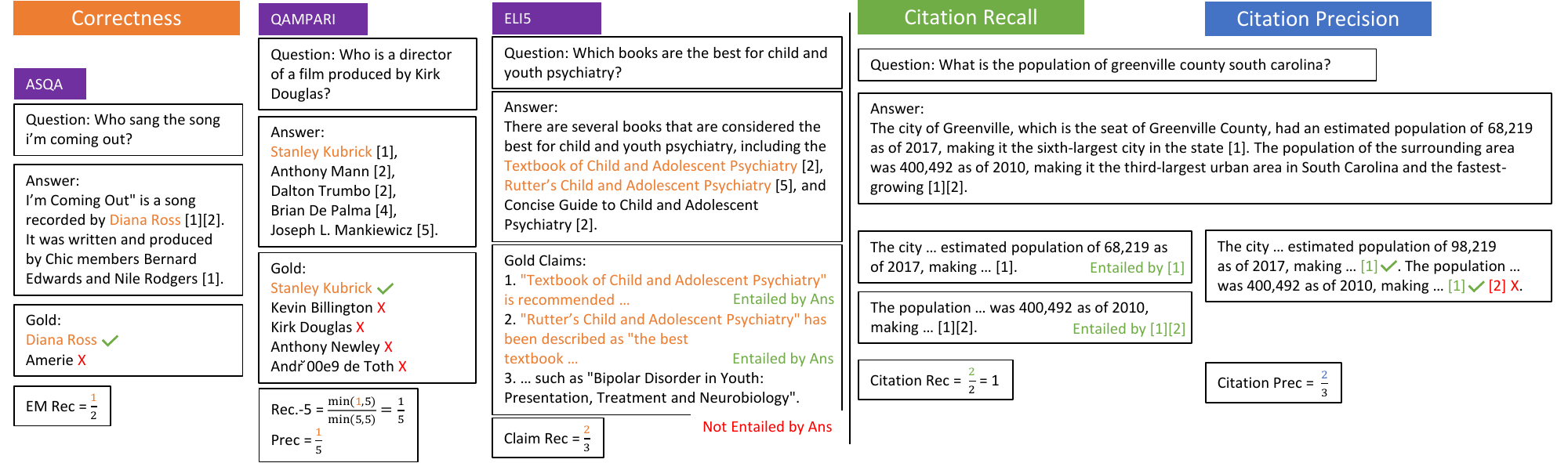}
    \caption{\label{fig:metric_diagram} Left: Examples of how the Correctness metrics are computed for ASQA (EM Rec), QAMPARI (Rec.-5, Prec), and ELI5 (Claim Rec) respectively; Right: An example of how the Citation Recall and Citation Precision are computed.} 
\end{figure*}

We use the metrics from ALCE \cite{gao2023enabling} to evaluate our model generations' correctness and citation quality. Their computations are illustrated in Figure~\ref{fig:metric_diagram}.

We illustrate their computations with a response $y$ with $l_2$ sentences and $l_3$ citations to a question with $t$ gold answers (key information pieces). See examples in Figure~\ref{fig:metric_diagram}.

\paragraph{Correctness Recall} Suppose $y$ captures $h$ out of $t$ key information pieces (i.e., they have a Correctness Recall reward of $+w_1$ (See Appendix~\ref{app:correctness_recall})), then its Correctness Recall is $\frac{h}{t}$. The detailed implementations vary among datasets (\textbf{EM Rec};\textbf{Rec.-5};\textbf{Claim Rec}).

More specifically, since the key information lists are available in ASQA and QAMPARI, we compute an exact string match recall \textbf{(EM Rec)} that checks how many key information pieces are an exact substring of generated responses. On top of that, following \cite{gao2023enabling}, we compute \textbf{Rec.-5} (i.e., $\frac{min(h, 5)}{min(t, 5)}$) on QAMPARI, so that the recall will be 100 if at least 5 correct answers are hit. On ELI5, we only have a single long-form answer for each question. Therefore, we follow \cite{gao2023enabling} to infer three sub-claims for each answer with text-davinci-003, and use them as the key information pieces. We then use an NLI model to check how many sub-claims are entailed by the model response \textbf{(Claim Rec)}.

Additionally, on QAMPARI, we directly ask the model to format its response as a list of items. To check whether the model is blindly making more guesses to increase Correctness Recall, we include a \textbf{Correctness Precision} metric on QAMPARI. It measures the percentage of predicted items that match a key information piece. Suppose there are $t'$ comma-separated entries in the generated response, then the Correctness Precision will be $\frac{h}{t'}$.

\paragraph{Citation Recall} Suppose $l_2'$ out of $l_2$ sentences are entailed by the cited passages (i.e., they each have a Citation Recall reward of $+w_2$ (See Appendix~\ref{app:citation_recall})), then the Citation Recall is $\frac{l_2'}{l_2}$.

On QAMPARI, since the model generates a comma-separated list instead of a natural language response, we treat each comma-separated item (formatted as '\{question\} \{item\}') as a sentence.

\paragraph{Citation Precision} Suppose $l_3'$ out of $l_3$ citations are "relevant", (i.e., they each have a Citation Precision reward of $+w_3$ (See Appendix \ref{app:citation_precision})), then the Citation Precision is $\frac{l_3'}{l_3}$.

\paragraph{MAUVE} MAUVE \cite{pillutla2021mauve} is an additional metric in ALCE that is used to evaluate fluency, which we do not include in the main analysis since it is sensitive to output length and unstable \cite{gao2023enabling}. We only report it in Appenidx \ref{app:main_complete}. It is computed on the model's generated responses against the provided gold natural language responses. Following \cite{gao2023enabling}, we remove citations in the model's generations, append the question to both the gold and generated responses, and truncate them to the first 100 words before computing MAUVE.

\section{Additional Training Details}
\label{app:training}

\paragraph{Efficiency} For memory efficiency, we fine-tune LoRA \cite{hu2021lora} in the distillation and RS fine-tuning steps, and QLoRA \cite{dettmers2023qlora} for RLHF. Moreover, in RLHF training, we use an architecture similar to Hydra-PPO \cite{santacroce2023efficient} with a slight modification that separates the weights of the reward model from the main model. We use vLLM \cite{kwon2023efficient} to accelerate inference.

\paragraph{Context and generation length} We set the maximum source length to be 1,024 for T5-large\footnote{In addition to LLaMA-2-7B, we also apply our method to T5-large. Its results are in Appendix~\ref{app:main_complete}.} and 1,200 for LLaMA-2-7B since its prompts contains instructions. We set the maximum generation length to 200 for both models to control the conciseness of answers.

\paragraph{Distillation} We instruction-tune our model for 10 epochs with Adam Optimizer. We use a learning rate of 5e-5 and a linear learning rate scheduler for T5-large. We use a learning rate of 2e-4 and a cosine learning rate scheduler with a warm-up ratio of 0.03 for LLaMA-2-7B. The best checkpoints are selected using the dev set for RS fine-tuning or RL.

\paragraph{RS fine-tuning} When fine-tuning the model, we use the same hyperparameters as in the distillation step, except that we change the number of training epochs to 3.

\paragraph{LoRA} We set the rank to 64 and $\alpha$ to 16. The adapters are applied to the Q and V projection layers in the transformer unit. We use the same set of hyperparameters for every adapter.

\paragraph{RL} Similar to \citet{wu2023fine}, we set $\gamma=1$ and $\lambda=0.95$. We use Adam Optimizer. The learning rate is set to 1e-5 for T5-large and 3e-5 for LLaMA-2-7B, and a linear scheduler is used for both models. Additionally, we sample 4 response sequences from the model for each training sample, with the temperature set to 0.7 and top k set to 20. On T5-large, we train the model for 80,000 episodes in total. On LLaMA-2-7B, for efficiency reasons, we set the number of episodes to 48,000. We use 100 warm-up steps across every setting. The KL coefficient $\beta$ is set to 0.3. We repeat each RL experiment 3 times, each with a different seed. The results are the average of the inference results of these 3 independent runs.

\paragraph{NLI model} For reward computation, we use TRUE\footnote{See \url{https://huggingface.co/google/t5_xxl_true_nli_mixture}.} \cite{honovich2022true}, a T5-11B model for NLI, to measure entailment for correctness recall on ELI5, and citation recall and precision across our datasets. 
For efficiency, we replace TRUE with a smaller NLI model, namely attrscore-flan-t5-xl \cite{yue2023automatic}, during the RL training stage.

\section{Complete Main Experiments}
\label{app:main_complete}

\begin{table*}[!htp]
	\centering
    \small
	\resizebox{1.0\textwidth}{!}{
    \begin{tabular}{lcccc|cccc|cccc}
		\toprule
        & \mc{4}{c}{ASQA} & \mc{4}{c}{QAMPARI} & \mc{4}{c}{ELI5} \\
	& \mc{1}{c}{Correctness} & \mc{2}{c}{Citation} & \mc{1}{c}{Fluency} & \mc{2}{c}{Correctness} & \mc{2}{c}{Citation} & \mc{1}{c}{Correctness} & \mc{2}{c}{Citation} & \mc{1}{c}{Fluency}\\
		\cmidrule{2-5} \cmidrule{6-9} \cmidrule{10-13}
	  & EM Rec & Rec & Prec & MAUVE & Rec.-5 & Prec & Rec & Prec & Claim Rec & Rec & Prec & MAUVE\\
		\midrule
        ICL {\scriptsize ChatGPT} & 39.96 & 74.72 & 70.97 & 76.49 & 18.34 & 18.57 & 17.65 & 20.60 & \textbf{13.47} &50.94 &47.58 &43.13\\
        ICL {\scriptsize LLaMA-2-7B} & 34.15 & 14.12 & 15.26 & 85.31 &8.24&9.23&7.51&7.65&7.83&14.44&8.92&43.53\\
        ICL {\scriptsize LLaMA-2-13B} & 35.97 & 21.59 & 25.27 & 88.01 &12.66&12.88&8.34&8.42&13.20&6.86&4.51&41.15\\
        Self-RAG$_{7B}$* & 30.0 & 67.8 & 66.9 & 74.3 &-&-&-&-&-&-&-&-\\
        Self-RAG$_{13B}$* & 31.7 & 71.3 & 70.3 & 71.6 &-&-&-&-&-&-&-&-\\
        \rowcolor{lightgray}\mc{13}{c}{\textit{T5 (Separate Setting)}}\\
        $\mathcal{M}_{dist}$ &  33.60  &    71.98 &  65.15  & 66.81 &15.98& 14.30&15.48&17.96&9.03&48.32&43.29&\colorbox{yellow}{51.04} \\
	~~~+f.g.RL & 33.75 & 74.24 & 68.30 & 61.11 &18.10&14.56&15.75&17.97&9.63&\colorbox{yellow}{52.04}&47.25&40.31\\
        \rowcolor{lightgray}\mc{13}{c}{\textit{LLaMA-2-7B (Separate Setting)}}\\
        $\mathcal{M}_{dist}$ & 34.84 & 69.30 & 62.20 & 59.93 &17.66&16.42&16.14&17.93&12.77&49.37&43.73&\colorbox{yellow}{48.53}\\
        ~~~+h.RL & 36.80 & 71.57 & 68.87 & \colorbox{yellow}{77.79} &17.30&17.08&16.80&18.56&12.40&49.95&45.83&\colorbox{yellow}{54.68}\\
        ~~~+f.g.RL & 36.94 & 71.86 & 69.75 & \colorbox{yellow}{77.21} &\colorbox{yellow}{\textbf{18.86}}&17.66&16.95&18.36&12.23&49.76&47.10&\colorbox{yellow}{47.23}\\
        ~~~+h.RS & 37.41 & \colorbox{yellow}{77.84} & \colorbox{yellow}{71.28} & 68.71 &16.06&18.01&17.29&18.13&11.30&\colorbox{yellow}{55.93}&\colorbox{yellow}{51.35}&42.75\\
        ~~~+f.g.RS & \colorbox{yellow}{40.24} & \colorbox{yellow}{77.65} & \colorbox{yellow}{74.96} & \colorbox{yellow}{83.10} &17.48&\colorbox{yellow}{20.67}&\colorbox{yellow}{20.62}&\colorbox{yellow}{21.65}&11.87&\colorbox{yellow}{61.27}&\colorbox{yellow}{56.45}&\colorbox{yellow}{53.37}\\
        ~~~+h.(RS+RL)  & 37.72 & \colorbox{yellow}{78.47} & \colorbox{yellow}{73.19} & 68.65 &17.58&\colorbox{yellow}{20.74}&\colorbox{yellow}{21.25}&\colorbox{yellow}{21.61}&11.30&\colorbox{yellow}{56.67}&\colorbox{yellow}{53.69}&\colorbox{yellow}{\textbf{59.79}}\\
        ~~~+f.g.(RS+RL) & \colorbox{yellow}{\textbf{41.33}} & \colorbox{yellow}{\textbf{79.71}} & \colorbox{yellow}{\textbf{77.90}} & \colorbox{yellow}{\textbf{91.14}}&15.76&\colorbox{yellow}{\textbf{21.60}}&\colorbox{yellow}{\textbf{21.36}}&\colorbox{yellow}{\textbf{21.92}}&11.47&\colorbox{yellow}{\textbf{63.45}}&\colorbox{yellow}{59.96}&\colorbox{yellow}{51.32}\\
        \rowcolor{lightgray}\mc{13}{c}{\textit{LLaMA-2-7B (Combined Setting)}}\\
        $\mathcal{M}_{dist}$ & 35.56 & \colorbox{yellow}{74.80} & 67.99 & 62.08 &17.26&16.18&\colorbox{yellow}{18.69}&18.94&12.03&49.69&45.71&36.03\\
        ~~~+h.RL & 34.33 & \colorbox{yellow}{75.77} & 70.12 & 61.79 & 17.30 & 16.44 & 16.39 & 18.92 & 11.52 & \colorbox{yellow}{51.77} & \colorbox{yellow}{49.32} & \colorbox{yellow}{45.46} \\
        ~~~+f.g.RL & 35.99 & \colorbox{yellow}{76.30} & \colorbox{yellow}{72.38} & 69.99 &\colorbox{yellow}{18.39}&\colorbox{yellow}{18.81}&\colorbox{yellow}{17.82}&19.07&11.60&\colorbox{yellow}{51.29}&\colorbox{yellow}{51.09}&40.64\\
        ~~~+h.RS & 37.96 & \colorbox{yellow}{74.86} & 68.48 & \colorbox{yellow}{79.04} &14.62&15.21&16.71&17.98&11.60&\colorbox{yellow}{54.10}&\colorbox{yellow}{48.95}&47.91\\
        ~~~+f.g.RS & \colorbox{yellow}{40.07} & \colorbox{yellow}{76.71} & \colorbox{yellow}{74.35} & \colorbox{yellow}{87.04} &16.14&\colorbox{yellow}{18.95}&\colorbox{yellow}{18.56}&19.50&11.67&\colorbox{yellow}{58.75}&\colorbox{yellow}{55.03}&\colorbox{yellow}{45.03}\\
        ~~~+h.(RS+RL) & 37.33 & \colorbox{yellow}{74.86} & 69.37 & \colorbox{yellow}{81.91} & 15.02&15.67&16.82&18.09&11.21&\colorbox{yellow}{55.62}&\colorbox{yellow}{50.58}&\colorbox{yellow}{47.76}\\
        ~~~+f.g.(RS+RL) & \colorbox{yellow}{40.05} & \colorbox{yellow}{77.83} & \colorbox{yellow}{76.33} & \colorbox{yellow}{89.08} & 16.65 &\colorbox{yellow}{19.54}&\colorbox{yellow}{19.50}&20.03&11.54&\colorbox{yellow}{60.86}&\colorbox{yellow}{\textbf{60.23}}&40.52\\
		\bottomrule
	\end{tabular}}
	\caption{\label{tab:main_complete_result} Complete main result on the three QA datasets. In-Context: In-Context Learning; * indicates results that are reported by the original paper; - indicates numbers not reported by the original paper; Entries that are highlighted in \colorbox{yellow}{yellow} are those that surpass their counterparts from ChatGPT. Our results on ChatGPT slightly deviate from those reported by \citet{gao2023enabling}, since they use sampling when decoding the responses during inference, and their evaluation on QAMPARI and ELI5 is run on the development split, while ours is run on the test split.}
\end{table*}

To demonstrate the general applicability of our framework, we also apply our method to T5-large \cite{raffel2020t5}. We provide the complete main experiment results in Table~\ref{tab:main_complete_result}, including results on T5-large and on LLaMA-2-7B checkpoints. These new models are trained in the $separate$ setting, which means they are trained on each individual dataset instead of the mixture. We also include additional baselines ICL {\scriptsize LLaMA-2-13B} and Self-RAG \cite{asai2023selfrag}. Note that RL experiments for the $separate$ training settings are run only once, since we observe little variation when repeating them with different seeds.

\section{Retrieval Analysis}
\label{app:retrieval_analysis}
\begin{table}[!htp]
	\centering
    \small
	\setlength{\tabcolsep}{2pt}
	\begin{tabular}{lc|ccc|c}
		\toprule
        & \mc{1}{c}{ASQA} & \mc{3}{c}{QAMPARI} & \mc{1}{c}{ELI5} \\
	& \mc{1}{c}{P.R.: 57.65} & \mc{3}{c}{P.R.: 20.53} & \mc{1}{c}{P.R.: 9.13}\\
	\cmidrule{2-2} \cmidrule{3-5} \cmidrule{6-6}
	  & C.R. & C.R.-5 & C.R. & C.P. & C.R.\\
		\midrule
        ICL {\scriptsize ChatGPT} & 73.05 & 68.81 & 67.57 & 16.86 & \textbf{93.28}\\
        ICL {\scriptsize LLaMA-2-7B} & 65.24  &    55.16 &  54.59  & 8.99 &91.15\\
        \midrule
        $\mathcal{M}_{dist}$ & 67.68 & 69.95 & 68.78 & 15.59 &92.58\\
        \rowcolor{lightgray}\mc{6}{c}{\textit{RL/RS only}}\\
        ~~~+h.RL & 66.10 & 69.89 & 68.73 & 15.85 & 92.47\\
        ~~~+f.g.RL & 68.62 & \textbf{71.19} & \textbf{69.84} & 18.02 &92.34\\
        ~~~+h.RS & 70.87 & 63.90 & 63.03 & 14.74 & 92.68\\
        ~~~+f.g.RS & \textbf{74.28} & 67.27 & 66.16 & 18.41 & 92.73\\
        \rowcolor{lightgray}\mc{6}{c}{\textit{RS+RL}}\\
        ~~~+h.(RS+RL) & 70.19 & 64.67 & 63.79 & 15.25 & 92.41\\
        ~~~+f.g.(RS+RL) & 74.08 & 67.87 & 66.75 & \textbf{18.87} & 92.57\\
		\bottomrule
	\end{tabular}
	\caption{\label{tab:document} Retrieval Analysis. P.R.: Passage Recall (i.e., Correctness Recall@5 of the retrieved passages); C.R.: Correctness Recall calibrated with the answers in the retrieved passages; C.R.-5: Correctness Recall that is 100 if there are at least 5 hits, and calibrated with the retrieved passages; C.P.: Correctness Precision calibrated with the retrieved passages. 
    }
\end{table}

As can be seen from Table~\ref{tab:document}, around 70\% of answers present in the retrieved documents can be captured by our LLMs on ASQA and QAMPARI. On ELI5, this coverage increases to more than 90\%.

\section{Training Curves}

\begin{figure*}[ht!]
    \includegraphics[width=.33\textwidth]{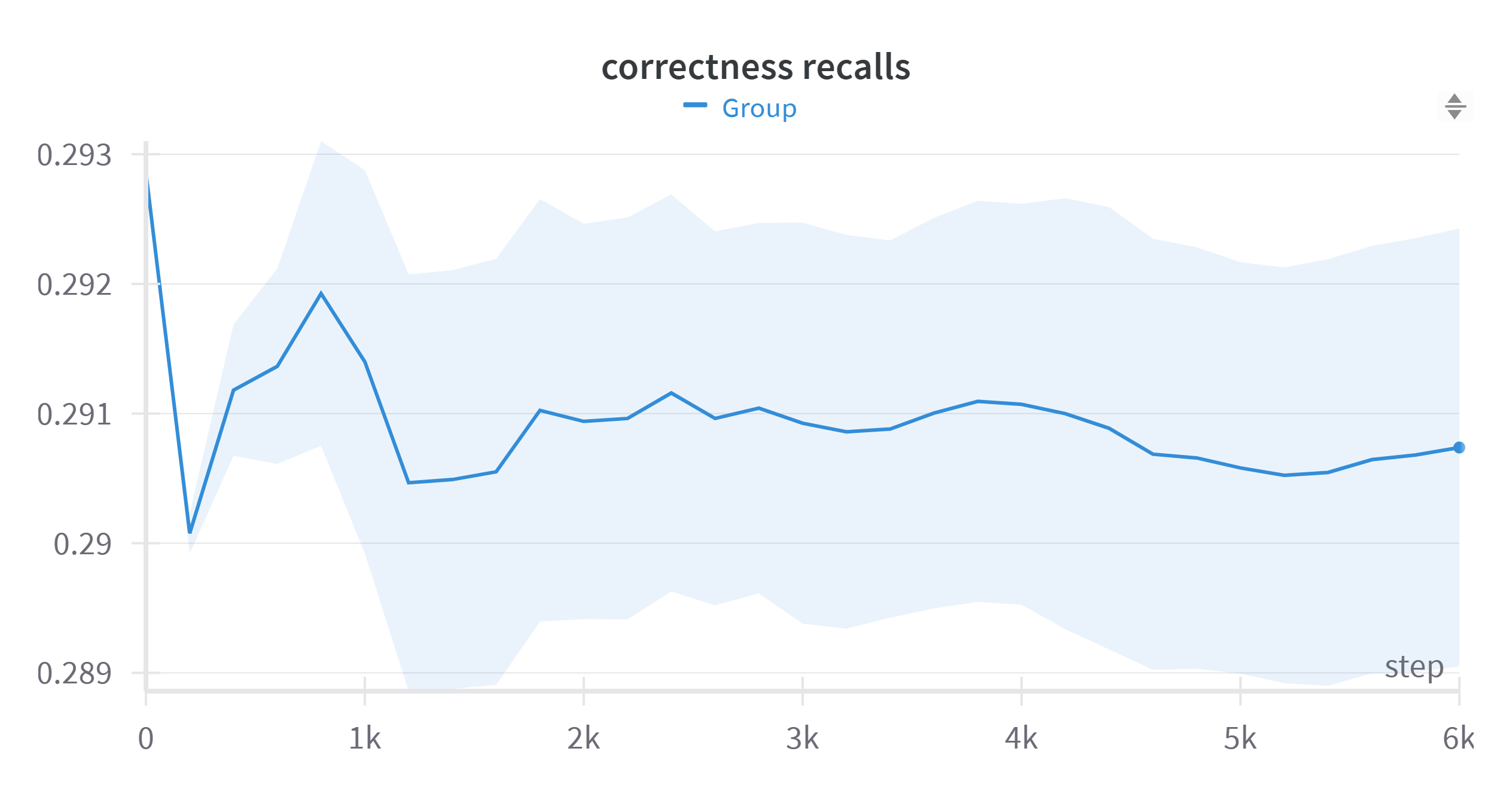}\hfill
    \includegraphics[width=.33\textwidth]{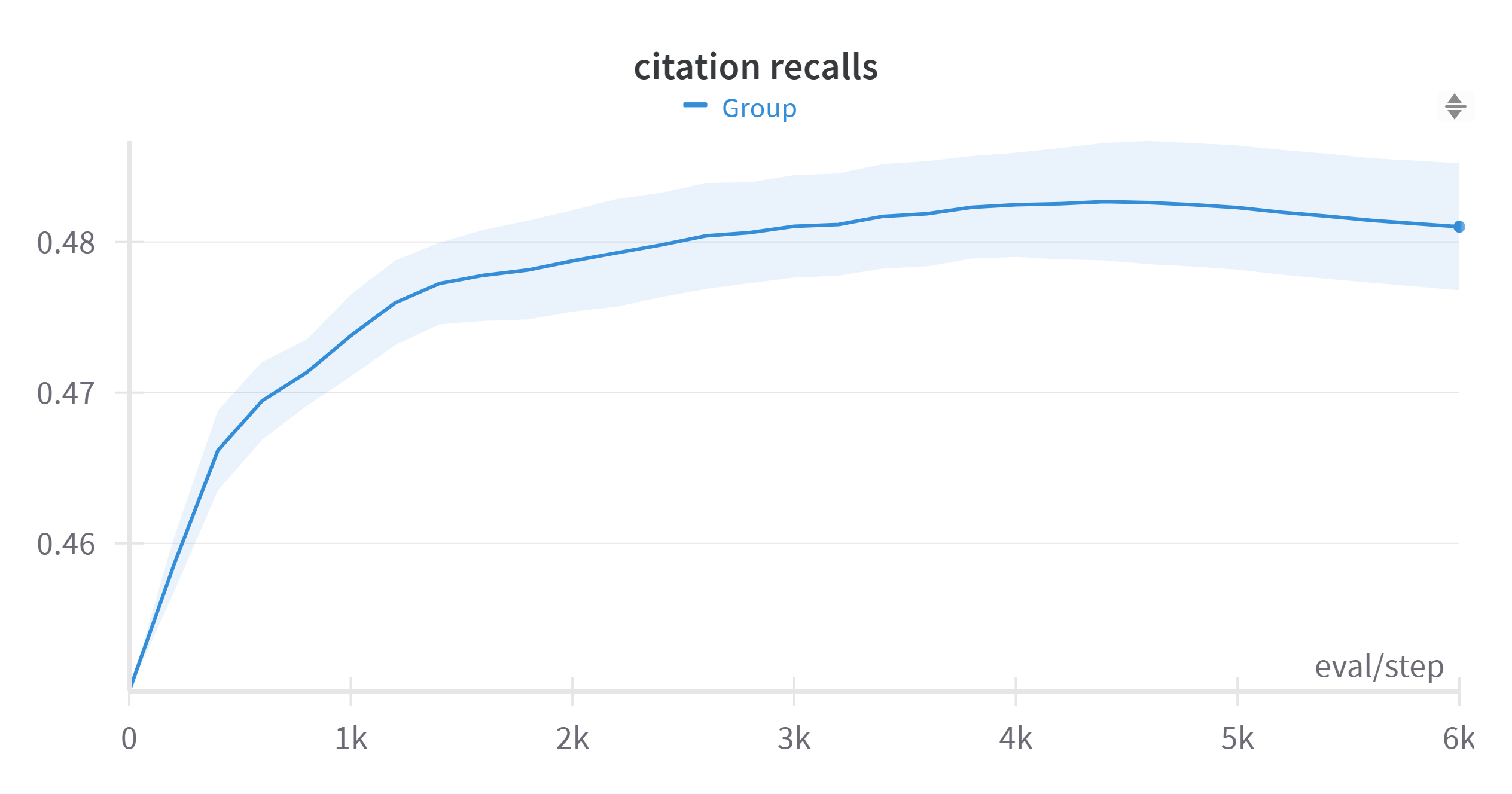}\hfill
    \includegraphics[width=.33\textwidth]{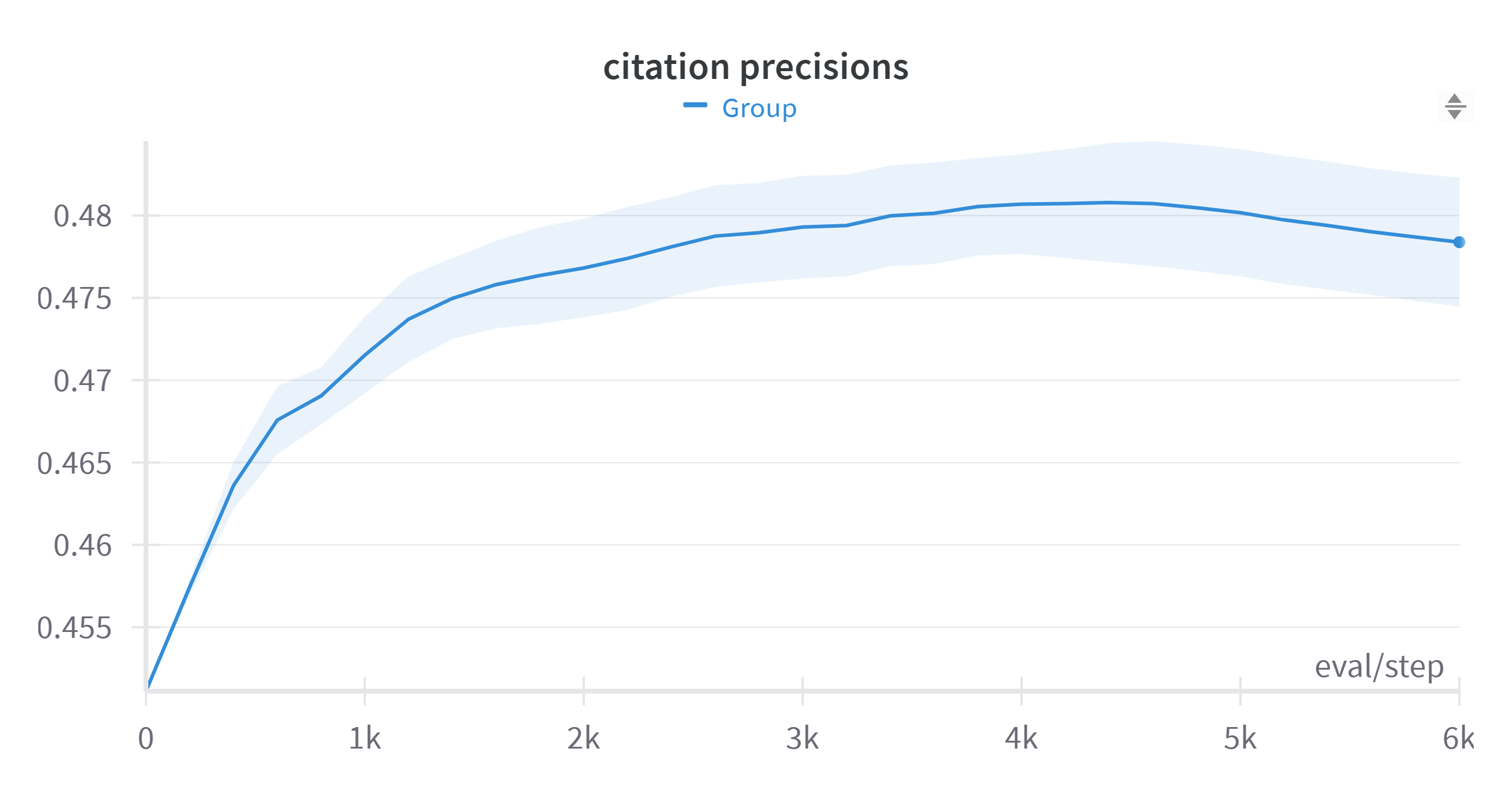}
    \caption{\label{fig:training_curves}Training curves of LLaMA-2-7B with f.g.RL in the $combined$ setting, measured on the development set across 3 independent runs. The shaded region indicates the standard error across these runs.}
\end{figure*}

We show the f.g.RL training curves of LLaMA-2-7B across the 3 independent runs in figure~\ref{fig:training_curves}. While correctness recall fluctuates, there is a steady increase in citation recall and citation precision.

\section{Citation Error Analysis}
\label{app:error_analysis_examples}
\begin{table}[!htp]
    \small
    \hrule
    Instruction: Provide a $\cdots$\\\\
    Question: Who is a director of a film produced by Kirk Douglas?\\\\
    Document [1](Title: Kirk Douglas): began producing films as varied as "Paths of Glory" (1957) and $\cdots$\\
    Document [2](Title: Kirk Douglas): executive producer as well, raising the \$12 million production cost, making $\cdots$ Douglas initially selected Anthony Mann to direct, but replaced him early on with \textcolor{Tan}{Stanley Kubrick}, with whom $\cdots$ When the film was released, Douglas gave full credit to its screenwriter, Dalton Trumbo, who was on the $\cdots$\\
    Document [3] $\cdots$\\\\
    Answers: \textcolor{Tan}{Stanley Kubrick [1]}, Anthony Mann [2], Dalton Trumbo [2], Brian De Palma [4], Joseph L. Mankiewicz [5].\\
    \hrule
    \caption{\label{tab:citation_error_1} An example from QAMPARI test set demonstrating the mixing up of passage ids error. \textcolor{Tan}{Stanley Kubrick} is only mentioned in [2], but is erroneously attributed to [1].}
\end{table}

\begin{table}[!htp]
    \small
    \hrule
    Instruction: Write an $\cdots$\\\\
    Question: What exactly is ’spin’ in quantum mechanics?\\\\
    Document [1](Title: quantum mechanics - Does spin have anything to do with a rate of change? - Physics Stack Exchange): quantum mechanics - Does spin have anything to do $\cdots$ what exactly does it mean, 1/2\\
    Document [2](Title: quantum mechanics - Does spin have anything to do with a rate of change? - Physics Stack Exchange): quantum mechanics -Does spin have anything to do $\cdots$ why do we relate spin with angular momentum operator in quantum mechanics?\\
    Document [3] $\cdots$\\\\
    Answers: In quantum mechanics, spin refers to the intrinsic angular momentum of particles, such as electrons, which can have either \textcolor{olive}{a spin up or a spin down state [1][2]}. Spin is $\cdots$ \\
    \hrule
    \caption{\label{tab:citation_error_2} An example from ELI5 test set demonstrating the redundant citations error. None of the documents from [1] to [5] mention anything about \textcolor{olive}{spin states}. The LLM retrieves its parametric knowledge when answering the question, but it still inserts the citations.}
\end{table}

\begin{table}[!htp]
    \small
    \hrule
    Instruction: Provide a $\cdots$\\\\
    Question: Michael Kiske is part of what \textcolor{YellowOrange}{music group}?\\\\
    Document [1] $\cdots$\\
    Document [2] $\cdots$\\
    Document [3](Michael Kiske): $\cdots$ In 2006, Michael Kiske released \textcolor{YellowOrange}{his third full-length solo album under the name of "Kiske"}. The album contained songs in a mellow, acoustic singer-songwriter style and was a step away from the sound of his previous solo albums. $\cdots$\\
    Document [4] $\cdots$\\\\
    Answers: Helloween [1], Unisonic [1], SupaRed [3], \textcolor{YellowOrange}{Kiske [3]}, Ill Prophecy [5].\\
    \hrule
    \caption{\label{tab:citation_error_3} An example from QAMPARI test set demonstrating the misinterpretation of a single passage error. \textcolor{YellowOrange}{"Kiske"} is actually the name of Michael Kiske's album, but is incorrectly interpreted by the LLM as the name of a \textcolor{YellowOrange}{music group} where Michael Kiske is a member.}
\end{table}

\begin{table}[!htp]
    \small
    \hrule
    Instruction: Write an $\cdots$\\\\
    Question: Purdue football players in nfl hall of fame?
\\\\
    Document [1] $\cdots$\\
    Document [2](Purdue Boilermakers football): $\cdots$ Among the numerous Boilermakers that have participated in the NFL, CFL, and AFL, a total of \textcolor{YellowOrange}{34} have received all-star recognition by their respective\\
    Document [3](Purdue Boilermakers football): $\cdots$ A total of seventeen Boilermakers have been inducted into the College Football Hall of Fame. \textcolor{YellowOrange}{Four Boilermakers} hold the distinguished title of Pro Football Hall of Fame inductees. $\cdots$\\
    Document [4] $\cdots$\\\\
    Answers: As of September 2018, there are a total of \textcolor{YellowOrange}{34 Purdue Boilermakers} who have received all-star recognition by their respective NFL teams [2]. Four of \textcolor{YellowOrange}{these} players have been inducted into the Pro Football Hall of Fame [3]. $\cdots$\\
    \hrule
    \caption{\label{tab:citation_error_4} An example from ASQA test set demonstrating the incorrect synthesis of multiple passages error. The coreference \textcolor{YellowOrange}{these} in LLM's response asserts that the 4 boilermakers who have been inducted into the Pro Football Hall of Fame are among the 34 boilermakers who have received all-star recognition, while such connection is not established by the original passages [2] and [3].}
\end{table}

We show examples for each type of citation error in Table~\ref{tab:citation_error_1}, \ref{tab:citation_error_2}, \ref{tab:citation_error_3}, and \ref{tab:citation_error_4}.

\section{Additional Experiment on EXPERTQA}
\label{app:expertqa}
\begin{table}[!htp]
	\centering
    \small
	\setlength{\tabcolsep}{2pt}
	\begin{tabular}{lcccc}
		\toprule
        & \mc{4}{c}{EXPERTQA} \\
        & \mc{1}{c}{Correctness} & \mc{2}{c}{Citation} & \mc{1}{c}{Fluency}\\
		\cmidrule{2-2} \cmidrule{3-4} \cmidrule{5-5}
	  & Claim Rec & Rec & Prec & MAUVE \\
		\midrule
        ICL {\scriptsize ChatGPT} & \textbf{19.27} & 47.79 & 47.30 & \textbf{48.68}\\
        ICL {\scriptsize LLaMA-2-7B} & 10.32 & 10.09 & 7.79 & 34.27\\
        \midrule
        $\mathcal{M}_{dist}$ & 15.28 & 49.03& 46.22 & 40.63\\
        ~~~+f.g.RL & 15.53 & 49.73 & 51.11 & 45.92\\
        ~~~+f.g.RS & 15.48 & \textbf{59.46} & 57.58 & 44.67\\
        ~~~+f.g.(RS+RL) & 15.23 & 58.94 & \textbf{59.80} & 42.13\\
		\bottomrule
	\end{tabular}
	\caption{\label{tab:additional_experiment_expertqa} Result on EXPERTQA.}
\end{table}

We also evaluate EXPERTQA with our own metrics. The results are shown in Table~\ref{tab:additional_experiment_expertqa}. Similar to ELI5, our method outperforms ChatGPT on citation recall and precision, but still lags behind on claim recall. The reasons can be complicated, while the most probable ones are the same as those for ELI5, including (1) insufficient emphasis on correctness reward ($w_1$) during RS and RL and (2) incorrect sub-claims derivation by text-davinci-003 that can cause inconsistency between the rewards and true correctness recall of the generated responses.


\section{Prompts}
\label{app:prompts}

\begin{table*}[!htp]
    \rule{\textwidth}{1pt}
    Instruction: Write an accurate, engaging, and concise answer for the given question using only the provided search results (some of which might be irrelevant) and cite them properly. Use an unbiased and journalistic tone. Always cite for any factual claim. When citing several search results, use [1][2][3]. Cite at least one document and at most three documents in each sentence. If multiple documents support the sentence, only cite a minimum sufficient subset of the documents.\\
    
    Question: Who played galen in planet of the apes?\\
    
    Document [1](Title: Planet of the Apes): \{text 1\}\\
    Document [2](Title: Planet of the Apes (1968 film)): \{text 2\}\\
    Document [3](Title: Planet of the Apes (1968 film)): \{text 3\}\\
    Document [4](Title: Planet of the Apes): \{text 4\}\\
    Document [5](Title: Planet of the Apes): \{text 5\}\\
    
    Answer:In the 1968 film Planet of the Apes, Galen was played by Wright King [2]. And in the tv series Planet of the Apes, Galen was played by Roddy McDowall [1].\\\\

    Question: Which is the most rainy place on earth?\\
    
    Document [1](Title: Cherrapunji): \{text 1\}\\
    Document [2](Title: Cherrapunji): \{text 2\}\\
    Document [3](Title: Mawsynram): \{text 3\}\\
    Document [4](Title: Earth rainfall climatology): \{text 4\}\\
    Document [5](Title: Going to Extremes): \{text 5\}\\
    
    Answer:Several places on Earth claim to be the most rainy, such as Llor\u00f3, Colombia, which reported an average annual rainfall of 12,717 mm between 1952 and 1989, and L\u00f3pez de Micay, Colombia, which reported an annual 12,892 mm between 1960 and 2012 [3]. However, the official record is held by Mawsynram, India with an average annual rainfall of 11,872 mm [3], although nearby town Sohra, India, also known as Cherrapunji, holds the record for most rain in a calendar month for July 1861 and most rain in a year from August 1860 to July 1861 [1].\\\\

    Question: \{question\}\\
    
    Document [1](Title: {title 1}): \{text 1\}\\
    Document [2](Title: {title 2}): \{text 2\}\\
    Document [3](Title: {title 3}): \{text 3\}\\
    Document [4](Title: {title 4}): \{text 4\}\\
    Document [5](Title: {title 5}): \{text 5\}\\
    
    Answer:\\
    \rule{\textwidth}{1pt}
    \caption{\label{tab:asqa_prompt} Prompt and demonstrations for ASQA.}
\end{table*}

\begin{table*}[!htp]
    \rule{\textwidth}{1pt}
    Instruction: Provide a list of accurate answers for the given question using only the provided search results (some of which might be irrelevant) and cite them properly. Always cite one and only one document for each answer. Separate answers by commas. For questions that have more than 5 answers, write at least 5 answers.\\
    
    Question: Which film has Gong Li as a member of its cast?\\
    
    Document [1](Title: Gong Li): \{text 1\}\\
    Document [2](Title: Gong Li): \{text 2\}\\
    Document [3](Title: Gong Li): \{text 3\}\\
    Document [4](Title: Zhang Yimou): \{text 4\}\\
    Document [5](Title: Gong Li): \{text 5\}\\
    
    Answer:The Story of Qiu Ju [1], Farewell My Concubine [2], Flirting Scholar [2], The Monkey King 2 [3], Mulan [3], Saturday Fiction [3], Coming Home [3].\\\\

    Question: Glenn Ford was a member of cast in which film?\\
    
    Document [1](Title: Glenn Ford): \{text 1\}\\
    Document [2](Title: Glenn Ford): \{text 2\}\\
    Document [3](Title: CBS Thursday Night Movie): \{text 3\}\\
    Document [4](Title: The Trouble with Girls (film) ): \{text 4\}\\
    Document [5](Title: Trouble in the Glen): \{text 5\}\\
    
    Answer:So Ends Our Night [1], Heaven with a Barbed Wire Fence [1], Happy Birthday to Me [2], The Greatest Gift [2], The Gift [2], The Brotherhood of the Bell [3].\\\\

    Question: \{question\}\\
    
    Document [1](Title: {title 1}): \{text 1\}\\
    Document [2](Title: {title 2}): \{text 2\}\\
    Document [3](Title: {title 3}): \{text 3\}\\
    Document [4](Title: {title 4}): \{text 4\}\\
    Document [5](Title: {title 5}): \{text 5\}\\
    
    Answer:\\
    \rule{\textwidth}{1pt}
    \caption{\label{tab:qampari_prompt} Prompt and demonstrations for QAMPARI.}
\end{table*}

\begin{table*}[!htp]
    \rule{\textwidth}{1pt}
    Instruction: Write an accurate, engaging, and concise answer for the given question using only the provided search results (some of which might be irrelevant) and cite them properly. Use an unbiased and journalistic tone. Always cite for any factual claim. When citing several search results, use [1][2][3]. Cite at least one document and at most three documents in each sentence. If multiple documents support the sentence, only cite a minimum sufficient subset of the documents.\\
    
    Question: What's the difference between Shia vs. Sunni Islam?\\
    
    Document [1](Title: The Sunni vs Shia Divide - Explained - Globaloi): \{text 1\}\\
    Document [2](Title: What\u2019s the difference between Sunni and Shia Islam? \u2013 Macrosnaps): \{text 2\}\\
    Document [3](Title: Difference between Sunni and Shia Muslims | Sunni vs Shia Muslims): \{text 3\}\\
    Document [4](Title: What is the difference between Shia and Sunni Islam? - Islam Stack Exchange): \{text 4\}\\
    Document [5](Title: What is the difference between Sunni and Shia Islam? | Patrick Syder Travel): \{text 5\}\\
    
    Answer:The main difference between Shia and Sunni Muslim is related to ideological heritage and issues of leadership [1]. This difference is first formed after the death of the Prophet Muhammad in 632 A.D. [1][2]. The ideological practice of the Sunni branch strictly follows Prophet Muhammad and his teachings, while the Shia branch follows Prophet Muhammad's son-in-law Ali [2]. Nowadays, Sunni and Shia are the major branches of Islam [3].\\\\

    Question: How do student loans affect getting a mortgage?\\
    
    Document [1](Title: Student Loans \u2013 How do they work? | The Financial Review): \{text 1\}\\
    Document [2](Title: How Does Student Loan Debt Affect Buying a Home? | Experian): \{text 2\}\\
    Document [3](Title: Studentloanify - How your student loans affect your home mortgage prospects): \{text 3\}\\
    Document [4](Title: How do student loans affect your credit score? | Student Loan Planner): \{text 4\}\\
    Document [5](Title: Does Student Loan Debt Affect Getting A Mortgage?): \{text 5\}\\
    
    Answer:When applying for a mortgage, student loans can affect the debt to income ratio, which is a key factor in determining the amount that an individual can afford to pay for the mortgage [1]. While student loan repayments do not appear in an individual's credit history and do not affect credit scores, lenders do consider the amount of an individual's student loan repayments when assessing their mortgage application [1][2][3]. Some 83\% of non-homeowners say student loan debt is preventing them from buying a home, according to the National Association of Realtors [2]. It is important to note that student loans do not prevent an individual from getting a mortgage [1].\\\\

    Question: \{question\}\\
    
    Document [1](Title: {title 1}): \{text 1\}\\
    Document [2](Title: {title 2}): \{text 2\}\\
    Document [3](Title: {title 3}): \{text 3\}\\
    Document [4](Title: {title 4}): \{text 4\}\\
    Document [5](Title: {title 5}): \{text 5\}\\
    
    Answer:\\
    \rule{\textwidth}{1pt}
    \caption{\label{tab:eli5_prompt} Prompt and demonstrations for ELI5.}
\end{table*}

The prompts for ASQA, QAMPARI, and ELI5 are shown in Table~\ref{tab:asqa_prompt}, Table~\ref{tab:qampari_prompt}, and Table~\ref{tab:eli5_prompt} respectively. The prompt for EXPERTQA is the same as that for ELI5, except that we append an additional sentence 'If you don't know the answer, just say "I don't know" and explain why the context is insufficient to answer the question.' to the instruction, in order to explicitly allow the model to abstain from answering the question. The reason is that EXPERTQA questions usually require more intensive domain knowledge to answer. Our prompts are consistent with those used by \citet{gao2023enabling}. Note that the in-context demonstrations are included in the prompt only in the distillation step (\S~\ref{sec:distill}) and the ICL baselines.

\section{Examples}

\begin{table*}[!htp]
    \rule{\textwidth}{1pt}
    Instruction: Write an $\cdots$\\    

    Question: Who sang the song i'm coming out?\\
    
    Document [1](Title: I'm Coming Out): I'm Coming Out "I'm Coming Out" is a song recorded by American singer Diana Ross. It was written and produced by Chic members Bernard Edwards and Nile Rodgers, and released in August 22, 1980 as the second single from Ross' self-titled tenth album "Diana" (1980). In 1979, Ross commissioned Chic founders Nile Rodgers and Bernard Edwards to create material for a new album after taking her daughters to see the band in concert, following the success of Ross's final Ashford \& Simpson-composed and produced LP," The Boss. "Rodgers got the idea for "I'm Coming Out" after noticing three different drag\\
    Document [2](Title: I'm Coming Out): the Loren Dawson remix. The original single version (which is similar to the original) can be found on both CD singles and the original motion picture soundtrack to "Maid in Manhattan". Ross' version appears on the film's soundtrack album as well. A music video was produced to promote the single. I'm Coming Out "I'm Coming Out" is a song recorded by American singer Diana Ross. It was written and produced by Chic members Bernard Edwards and Nile Rodgers, and released in August 22, 1980 as the second single from Ross' self-titled tenth album "Diana" (1980). In 1979, Ross commissioned Chic\\
    Document [3](Title: I'm Coming Out): "I'm Coming Out" and its resulting popularity has led to Ross's song often being sampled, most notably by Stevie J, who sampled the song for rapper The Notorious B.I.G.'s 1997 song "Mo Money Mo Problems". There's also been a cover by Marcia Hines. It was featured as a cover for the video game "Karaoke Revolution" as well. In more recent times, the song was sampled in Keyshia Cole's remake of "I'm Coming Out" (featuring Iggy Azalea) for the original soundtrack of the 2014 movie The Other Woman, starring Cameron Diaz, Kate Upton and Leslie Mann. The song is sampled in\\
    Document [4](Title: I'm Coming Out): "Break Your Heart Right Back" from Ariana Grande's sophomore album "My Everything" (2014). This song was also covered for the soundtrack to the 2016 movie "Trolls". In 1997, German comedian Hella von Sinnen sampled "I'm Coming Out" for her single "Mein Coming Out", using the term "coming out" in the context of coming out of the closet, or revealing one's homosexuality. In 2003, American singer Amerie covered "I'm Coming Out" for the soundtrack to the film "Maid in Manhattan". Her version was released as a single in select European countries and Australia. The final single version to be released was\\
    Document [5](Title: Coming Out (album)): "Poinciana (The Song of the Tree)" featured a solo by Michael Brecker. "Coming Out" debuted on Billboard's Top Pop Album chart on September 18, 1976 and peaked at No. 48. The album was re-released on CD in March 1993. Manhattan Transfer Musicians Production Coming Out (album) Coming Out is the third album by The Manhattan Transfer, released August 19, 1976 on Atlantic Records. On the advice of their manager Aaron Russo, the group downplayed the "nostalgic" tenor of their previous breakthrough album in favor of more contemporary songs. However, the throwback single "Chanson d'Amour" became a breakout single in Europe\\
    
    Answer: \textcolor{blue}{"I'm Coming Out" is a song recorded by Diana Ross [1][2]. It was written and produced by Chic members Bernard Edwards and Nile Rodgers [1].}\\
    \textcolor{orange}{Gold: "I'm Coming Out" is a song recorded by American singer Diana Ross. It was written and produced by Chic members Bernard Edwards and Nile Rodgers, and released on August 22, 1980, as the second single from Ross' self-titled tenth album Diana (1980). In 2003, American singer Amerie covered "I'm Coming Out" for the soundtrack to the film "Maid in Manhattan". Her version was released as a single in select European countries and Australia.}\\
    \textcolor{red}{Gold Answers: [Diana Ross, Amerie]}\\
    \rule{\textwidth}{1pt}
    \caption{\label{tab:asqa_example} ASQA example}
\end{table*}

\begin{table*}[!htp]
    \rule{\textwidth}{1pt}
    Instruction: Provide a $\cdots$\\    

    Question: Who is a director of a film produced by Kirk Douglas?\\
    
    Document [1](Title: Kirk Douglas): began producing films as varied as "Paths of Glory" (1957) and "Spartacus" (1960). In those two films, he collaborated with the then-relatively-unknown director Stanley Kubrick taking lead roles in both films. Douglas has been praised for helping to break the Hollywood blacklist by having Dalton Trumbo write "Spartacus" with an official on-screen credit, although this has been disputed by others. He produced and starred in "Lonely Are the Brave" (1962), considered a classic, and "Seven Days in May" (1964), opposite Burt Lancaster, with whom he made seven films. In 1963, he starred in the Broadway play "One Flew Over the\\
    Document [2](Title: Kirk Douglas): executive producer as well, raising the \$12 million production cost, making it one of the most expensive films made up to that time. Douglas initially selected Anthony Mann to direct, but replaced him early on with Stanley Kubrick, with whom he previously collaborated in "Paths of Glory". When the film was released, Douglas gave full credit to its screenwriter, Dalton Trumbo, who was on the Hollywood blacklist, and thereby effectively ended it. About that event, he said, "I've made over 85 pictures, but the thing I'm most proud of is breaking the blacklist." However the film's producer Edward Lewis and\\
    Document [3](Title: Spartacus (film)): credibility. Trumbo was publicly given credit for two major films. Otto Preminger made public that Trumbo wrote the screenplay for his film "Exodus", and Kirk Douglas publicly announced that Trumbo was the screenwriter of "Spartacus". Further, President John F. Kennedy publicly ignored a demonstration organized by the American Legion and went to see the film. After David Lean turned down an offer to direct, "Spartacus" was to be directed by Anthony Mann, then best known for his Westerns such as "Winchester '73" and "The Naked Spur". Douglas fired Mann at the end of the first week of shooting, in which\\
    Document [4](Title: Home Movies (film)): directing films. Kirk Douglas plays a film instructor loosely modeled on the director himself. Keith Gordon is one of his pupils who films everything that happens at home. Many of the events that happen to Gordon's character were modeled on events from De Palma's own adolescence, particularly rivalry with a more favored brother, a mother prone to dramatic outbursts, and a philandering father. Home Movies (film) Home Movies is a 1979 independent film directed by Brian De Palma and starring Kirk Douglas, Nancy Allen, Vincent Gardenia, Keith Gordon, Theresa Saldana, and Gerrit Graham. De Palma had been teaching film at\\
    Document [5](Title: Kirk Douglas): in the picture," adding that "he's as bright a person as I've met in the acting profession." She says that his "pragmatic approach to acting" would later be a "philosophy that ended up rubbing off on me." Between 1970 and 2008, Douglas made nearly 40 movies and appeared on various television shows. He starred in a western, "There Was a Crooked Man..." (1970), alongside Henry Fonda. The film was produced and directed by Joseph L. Mankiewicz. In 1973, he directed his first film, "Scalawag". Also in 1973, Douglas appeared in a made-for-TV musical version of "Dr. Jekyll and Mr. Hyde".\\
    
    Answer: \textcolor{blue}{Stanley Kubrick [1], Anthony Mann [2], Dalton Trumbo [2], Brian De Palma [4], Joseph L. Mankiewicz [5].}\\
    \textcolor{orange}{Gold: Stanley Kubrick, Kevin Billington, Kirk Douglas, Anthony Newley, Andr\u00e9 de Toth}\\
    \textcolor{red}{Gold Answers: [[Stanley Kubrick], [Kevin Billington], [Kirk Douglas], [Anthony Newley], [Andr\u00e9 de Toth]]}\\
    \rule{\textwidth}{1pt}
    \caption{\label{tab:qampari_example} QAMPARI example}
\end{table*}

\begin{table*}[!htp]
    \scriptsize
    \rule{\textwidth}{1pt}
    Instruction: Write an $\cdots$\\    

    Question: What exactly is 'spin' in quantum mechanics?\\
    
    Document [1](Title: quantum mechanics - Does spin have anything to do with a rate of change? - Physics Stack Exchange): quantum mechanics - Does spin have anything to do with a rate of change? - Physics Stack Exchange the electron. Not the answer you're looking for? Browse other questions tagged quantum-mechanics operators quantum-spin rotation spinors or ask your own question. How do you rotate spin of an electron? Did the Goudsmit-Uhlenbeck analysis of spin consider relativity? Is there an Ehrenfest-like result for the expectation value of orbital angular momentum? Can spin be related to a shift in angle? What does spin 0 mean exactly? Does String Theory explain spin? How can it be derived that particles described by the Dirac equation must have spin 1/2? When we say electron spin is 1/2, what exactly does it mean, 1/2\\
    Document [2](Title: quantum mechanics - Does spin have anything to do with a rate of change? - Physics Stack Exchange): quantum mechanics - Does spin have anything to do with a rate of change? - Physics Stack Exchange your own question. Did the Goudsmit-Uhlenbeck analysis of spin consider relativity? Can spin be related to a shift in angle? What does spin 0 mean exactly? Does String Theory explain spin? How can it be derived that particles described by the Dirac equation must have spin 1/2? When we say electron spin is 1/2, what exactly does it mean, 1/2 of what? Determine spin-state with $B=B_x$ in the $S_z$ base why do we relate spin with angular momentum operator in quantum mechanics?\\
    Document [3](Title: Search results for `Biology` - PhilArchive): Search results for `Biology` - PhilArchive biased to satisfy the collective goals of the ensemble. Using this postulate, we mathematically derive the correct spin probability distribution without using quantum mechanical formalism (operators and Born\u2019s rule) and exactly reproduce the quantum mechanical spin correlation in entangled pairs. Using our concept, we also mathematically derive the form of quantum mechanical wave function of free particle which is conventionally a postulate of quantum mechanics. Thus, we prove that the origin of quantum mechanical results lies in the will (or consciousness) of the objects biased by the collective goal of ensemble or universe. This biasing by the group on individuals\\
    Document [4](Title: The Upside to Quantum Mechanics): The Upside to Quantum Mechanics Questions About Quantum Mechanics Which is named Quantum Tunneling. The quantum simulator is installed as part of the QDK. There\u2019s no macroscopic property that\u2019s equivalent to quantum spin. The traditional manner of operating the Hopfield network is to continue picking neurons at random and updating them by thinking about the connected neurons, together with their weights. Basically, the concept is that, when you have a look at a system (such as an electron that\u2019s spinning in two directions at exactly the same time) the electron custom essays is going to be made to quit spinning in two directions simultaneously, pick\\
    Document [5](quantum mechanics - How exactly is fusion possible? - Physics Stack Exchange): quantum mechanics - How exactly is fusion possible? - Physics Stack Exchange Pauli's Exclusion Principle Can two electrons have the same momentum and spin directions? Nature of the quantum degeneracy pressure How does the Pauli-exclusion principle work if space is infinitely divisible? What exactly is meant by a quantum state in QM? Why a half-integer spin implies \u201cantisymmetry\u201d on the wave function? What is the physical (i.e. non-mathematical) cause off the difference between a system of identical bosons and fermions? How do two exactly the same (in space) wavefunctions of two non-distiguishable free particles contribute to Pauli's exclusion principle? How do I derive Pauli's exclusion principle with path integrals?\\
    
    Answer: \textcolor{blue}{In quantum mechanics, spin refers to the intrinsic angular momentum of particles, such as electrons, which can have either a spin up or a spin down state [1][2]. Spin is not related to a rate of change, but rather is a fundamental property of particles that can be described by quantum mechanical operators and the Pauli exclusion principle [1][5]. While there are various interpretations of quantum mechanics, including the idea that particles described by the Dirac equation must have spin 1/2 [2], the origin of quantum mechanical results is not necessarily related to the will or consciousness of the objects, as suggested in document [3].}\\
    \textcolor{orange}{Gold: Spin is the quantum mechanical analogue to rotational angular momentum. Rotational angular momentum is a measure of how strongly an object is rotating on its own axis. The Earth's revolution about its own axis (the one that causes days) is an example of rotational angular momentum.\\\\
    The reason I said that spin is an "analogue" to rotational angular momentum is that it's not an idea that really makes sense in QM. In order for us to see how an object is rotating, we have to be able to clearly see what shape it is. If you can't see a shape, rotation is sort of meaningless, since... well, you can't actually tell if something is rotating if it has no shape.\\\\
    Quantum mechanical objects don't have well-defined shapes. They're sort of fuzzy and nebulous, so you can't really pin them down like you'd want. We hence can't look at, say, an electron and go "oh, it's rotated by 180 degrees after two seconds, so its angular momentum is X." Electrons can't really rotate, it doesn't make sense.\\\\
    However, electrons do exhibit certain properties that make it *seem like* they're rotating. For example, when a charged (non-quantum mechanical object) rotates, it makes a magnetic field. Electrons also make a magnetic field in the exact same way. So, they seem to be acting like they're rotating, even though we can't really pin down what that means exactly.\\\\
    We thus call it "spin." It's... like they're rotating, hence "spin," but it's understood by physicists to not mean the exact same thing.\\\\
    Spin is "quantized," which means that there are only certain values that it can have. If you measure an electron's spin, you will only either get \u045b/2 (clockwise), or -\u045b/2 (counterclockwise)\^~(1). \u045b is the "reduced [Planck constant](\_URL\_0\_)," which shows up a lot in physics.\\\\
    \^~(1. note that when I say "measure," what's implied is that we're choosing an axis upon which to measure. For example, if you wanted to measure the Earth's angular momentum, you would measure it along the Earth's axis. The axis you choose to measure on for a quantum mechanical object is irrelevant though.)}\\
    \textcolor{red}{Gold Answer Claims:\\
    Spin in quantum mechanics is analogous to rotational angular momentum.\\
    Spin is not a concept that makes sense in quantum mechanics as objects do not have well-defined shapes.\\
    \hspace{20mm}Spin is quantized and can only have certain values when measured.
    }\\
    \rule{\textwidth}{1pt}
    \caption{\label{tab:eli5_example} ELI5 example}
\end{table*}

\begin{table*}[!htp]
    \small
    \rule{\textwidth}{1pt}
    Instruction: Write an $\cdots$\\    

    Question: Which books are the best for child and youth psychiatry?\\
    
    Document [1](Title: job details - Consulting Psychiatrist - Department of Psychiatry, Ambulatory mental health program, BC Children's Hospital in Vancouver): job details - Consulting Psychiatrist - Department of Psychiatry, Ambulatory mental health program, BC Children's Hospital in Vancouver themes. Through the affiliation with the BCHRI, the Department has established research strengths in neuroimaging, genetics, metabolism and clinical research. The Department is actively involved in supporting best practice through multiple provincial networks for which it provides infrastructure support (Eating Disorders, Health Literacy, Youth Concurrent Disorders and Child \& Youth Mental Health and Substance Use), through its robust and award winning Child \& Youth Mental Health Literacy Program and Child Health BC. Interested candidates will have obtained fellowship standing in the Royal College of Physicians \& Surgeons of Canada in Psychiatry along with certification in Child and Adolescent Psychiatry subspecialty\\
    Document [2](Title: Textbook of Child and Adolescent Psychiatry - Google Books): Textbook of Child and Adolescent Psychiatry - Google Books the Journal of the American Academy of Child and Adolescent Psychiatry. She has published more than 60 articles, book chapters, and books, including Concise Guide to Child and Adolescent Psychiatry, and Helping Parents, Youth, and Teachers Understand Medications for Behavioral and Emotional Problems, published by American Psychiatric Publishing, Inc. She is the principal author of the American Academy of Child and Adolescent Psychiatry Practice parameters for the assessment and treatment of children, adolescents, and adults with attention-deficit/hyperactivity disorder and a co-author of the American Academy of Child and Adolescent Psychiatry practice parameters for the use of stimulant medications. Title Textbook\\
    Document [3](Title: Journal of the American Academy of Child and Adolescent Psychiatry Page 5): Journal of the American Academy of Child and Adolescent Psychiatry Page 5 in evidence-based treatments for youth depression in recent decades, overall treatment effects are modest at best, with 30-50\% of youth being non-responders. Practice parameters consistently recommend systematic assessment and routine monitoring of depressive symptoms, or measurement-based care (MBC), to enhance youth depression treatment. However, the literature offers few guidelines on how to use assessment results to inform care decisions or detect real and clinically meaningful change. (Source: Journal of the American Academy of Child and Adolescent Psychiatry) Source: Journal of the American Academy of Child and Adolescent Psychiatry - February 27, 2019 Category: Psychiatry Authors: Freda F. Liu, Molly C.\\
    Document [4](Title: Browse In Child and Adolescent Psychiatry | Oxford Medicine Online - Oxford Medicine): Browse In Child and Adolescent Psychiatry | Oxford Medicine Online - Oxford Medicine Child and Adolescent Psychiatry x Download complete list of books in this Child and Adolescent Psychiatry Collection (.pdf) (.xls) Bipolar Disorder in Youth: Presentation, Treatment and Neurobiology Stephen M. Strakowski, Melissa P. DelBello, and Caleb M. Adler (eds) Bipolar disorders affect up to 3\% of the world's population and are the 6th leading cause of disability worldwide. As the disorder presents differently in youth as it does in adults, ... More Bipolar disorders affect up to 3\% of the world's population and are the 6th leading cause of disability worldwide. As the disorder presents differently in youth as it does\\
    Document [5](Title: Rutter's Child and Adolescent Psychiatry -ORCA): Rutter's Child and Adolescent Psychiatry -ORCA new look." -Professor Sir Michael Rutter, extract from Foreword. Reviews of previous editions: "This book is by far the best textbook of Child \& Adolescent Psychiatry written to date." -Dr Judith Rapoport, NIH "The editors and the authors are to be congratulated for providing us with such a high standard for a textbook on modern child psychiatry. I strongly recommend this book to every child psychiatrist who wants a reliable, up-to-date, comprehensive, informative and very useful textbook. To my mind this is the best book of its kind available today." -Journal of Child Psychology and Psychiatry\\
    
    Answer: \textcolor{blue}{There are several books that are considered the best for child and youth psychiatry, including the Textbook of Child and Adolescent Psychiatry [2], Rutter's Child and Adolescent Psychiatry [5], and Concise Guide to Child and Adolescent Psychiatry [2].}\\
    \textcolor{orange}{Gold: The books recommended for child and youth psychiatry include "Textbook of Child and Adolescent Psychiatry" which is noted in the Journal of the American Academy of Child and Adolescent Psychiatry  and "Rutter's Child and Adolescent Psychiatry", which was described as "the best textbook of Child \& Adolescent Psychiatry written to date" and a "very useful textbook" by Dr. Judith Rapoport at NIH and the Journal of Child Psychology and Psychiatry, respectively . Other resources could be found in the category of Child and Adolescent Psychiatry in Oxford Medicine Online, which includes a list of books such as "Bipolar Disorder in Youth: Presentation, Treatment and Neurobiology" .}\\
    \textcolor{red}{Gold Answer Claims:\\
        "Textbook of Child and Adolescent Psychiatry" is recommended for child and youth psychiatry.,\\
        "Rutter's Child and Adolescent Psychiatry" has been described as "the best textbook of Child \& Adolescent Psychiatry written to date".,\\
        Oxford Medicine Online has a list of books in the category of Child and Adolescent Psychiatry, such as "Bipolar Disorder in Youth: Presentation, Treatment and Neurobiology".
    }\\
    \rule{\textwidth}{1pt}
    \caption{\label{tab:expertqa_example} EXPERTQA example}
\end{table*}

We provide an example for ASQA, QAMPARI, ELI5, and EXPERTQA each, presented in Table~\ref{tab:asqa_example},~\ref{tab:qampari_example},~\ref{tab:eli5_example}, and~\ref{tab:expertqa_example} respectively. The responses are generated by the f.g.(RS + RL) checkpoints trained in the $combined$ setting.





\end{document}